\documentclass[runningheads]{llncs}

\usepackage{eccv}

\usepackage{eccvabbrv}

\usepackage{graphicx}
\usepackage{booktabs}

\usepackage{mathtools} %
\usepackage{graphicx}
\usepackage{tabularray}
\usepackage{tabularx}
\newcolumntype{C}{>{\centering\arraybackslash}X}
\usepackage{multirow}
\usepackage{microtype}
\usepackage{siunitx}
\usepackage{xparse}
\usepackage{fp}
\usepackage{float}
\usepackage[table]{xcolor}
\usepackage{wrapfig}
\usepackage{pifont} %
\newcommand{\cmark}{\ding{51}}

\definecolor{darkgreen}{RGB}{0,100,0}
\newcommand{\gss}[1]{\textcolor{darkgreen}{\scriptsize{#1}}}
\newcommand{\rss}[1]{\textcolor{red}{\scriptsize{#1}}}

\newcommand{\PAR}[1]{\vskip2pt \noindent{\bf #1~}}
\definecolor{riperow}{gray}{0.91}

\usepackage[accsupp]{axessibility}  %

\usepackage{hyperref}

\usepackage{orcidlink}

\begin{document}

\title{RIPE\texttt{++}: Reinforced Keypoint Learning from Positive Pairs Only}

\author{Johannes Künzel\inst{1,2}\orcidID{0000-0002-3561-2758} \and
Peter Eisert\inst{1,2}\orcidID{0000-0001-8378-4805} \and
Anna Hilsmann\inst{1}\orcidID{0000-0002-2086-0951}
}

\authorrunning{J. Künzel et al.}

\authorrunning{J. Künzel, P. Eisert, Anna Hilsmann}

\institute{Fraunhofer Heinrich-Hertz-Institute, HHI, Germany \and
Humboldt University Berlin, Germany
}

\maketitle

\begin{figure}[h]
    \centering
    \includegraphics[width=1.0\linewidth]{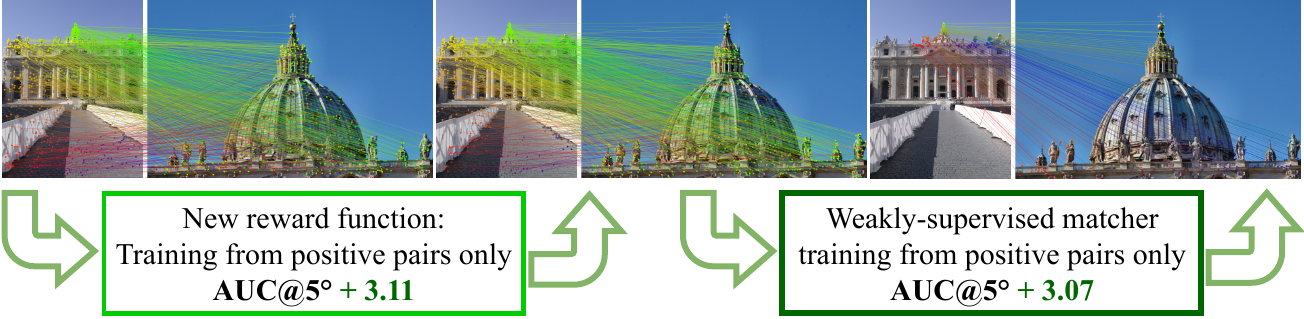}
    \caption{We propose a reward formulation for weakly-supervised keypoint extraction training that enables training from positive pairs only and is directly applicable to the training of learned matching architectures, demonstrated with LightGlue~\cite{lindenberger_2023_lightglue}.}
    \label{fig:teaser}
\end{figure}
\vspace{-0.8cm}

\begin{abstract}
Sparse keypoint extraction and matching underpin core tasks in geometric computer vision, including structure-from-motion, visual SLAM, augmented reality, and medical image registration. Learning robust local feature representations, however, typically requires accurate camera poses or depth supervision, which are often unavailable in real-world settings.
Reinforcement learning~(RL) has recently emerged as a promising alternative, requiring only the information if two images show the same scene or not.
However, existing RL formulations such as RIPE rely on coarse binary rewards and carefully constructed negative training pairs, limiting training stability and descriptor discriminability.
In this paper, we revisit RL-based keypoint learning and propose a reward that fully exploits the geometric consistency signal, deriving both reward and penalty from a single positive pair without contrasting against negatives.
This richer signal provides sufficient supervisory contrast to learn discriminative detectors and descriptors from positive image pairs alone, enabling representation learning under extremely limited supervision.
Furthermore, we show that the same RL objective can be extended to the matching stage by adapting LightGlue, raising AUC@5 on MegaDepth1500 from \num{56.58} to \num{59.65} and enabling weakly-supervised training of the full sparse matching pipeline from image pairs with partial visual overlap.
We validate our approach on established benchmarks, demonstrating competitive results compared to fully-supervised methods.
We further show that the method can be even trained on low texture medical video sequences, where camera poses are usually unavailable and standard SfM pipelines often fail.
Code and data are available at \url{https://github.com/fraunhoferhhi/RIPEpp}.
\end{abstract}
    
\section{Introduction}
\label{sec:introduction}

Sparse keypoint extraction and matching is a cornerstone of geometric computer vision, underpinning structure-from-motion (SfM), visual SLAM, augmented reality, and medical image registration.
Despite enormous progress, modern learned pipelines still depend on dense geometric supervision, i.e.\ ground-truth depth and camera poses, which presupposes calibrated hardware or offline SfM reconstruction and is expensive or impossible to acquire at scale.
The limiting factor in modern keypoint learning is thus no longer model capacity but the availability of geometric supervision, making representation learning under limited supervision a key challenge for geometric vision.

Reinforcement learning~(RL) circumvents the fact that the discrete nature of keypoint selection prevents direct gradient-based optimization and greatly reduces the required supervision, as shown in RIPE~\cite{knzel2025ripe-7fa}.
RIPE trained an RL-based detector and descriptor from image pairs annotated only with a binary same-scene/different-scene label, using RANSAC-based fundamental matrix estimation as a geometry-aware reward.
This reduces supervision to a single bit per pair, obtainable from the temporal structure of a video stream at no annotation cost.
Yet two limitations remain: the binary reward is coarse, limiting training stability and descriptor discriminability while requiring carefully constructed negative pairs; and the pipeline still depends on a matcher trained with full pose or depth supervision, negating much of the practical benefit.

We revisit RL-based keypoint learning and show that the effectiveness of weak supervision critically depends on how geometric consistency is translated into a reward signal.
We therefore propose a new reward based on the number of geometrically plausible inliers \emph{and} outliers.
This richer signal makes negative pairs unnecessary: the reward itself provides sufficient contrast to learn discriminative detectors and descriptors from positive pairs alone.
We further extend the RL paradigm to the matching stage by adapting LightGlue to the same weakly-supervised objective, raising the AUC@5 on MegaDepth1500 from \num{56.58} to \num{59.65}, and validate on established benchmarks and on medical video sequences, a domain where ground-truth poses are unavailable and standard SfM pipelines such as COLMAP cannot be applied.
Our method thus operates in the very-limited-resource regime along two axes, annotation (only binary) and data curation (raw video streams, no negative-pair mining), demonstrating that accurate local feature representations can be learned with minimal supervision and dataset preparation.
Our major contributions are as follows:
 
\begin{itemize}
    \item We address the limitations of the binary reward formulation in RIPE and propose a geometric reward that generates reward and penalty by taking inliers \emph{and} outliers into account, which provides a richer and more stable training signal for RL-based keypoint learning.
    \item This improved reward formulation naturally removes the need for negative training pairs, simplifying the training pipeline and dataset curation.
    \item We show that this allows training from data sources as simple as video streams, as we show for medical image data.
    \item We extend the same reward to a transformer-based matcher (LightGlue), improving AUC@5 on MegaDepth1500 from \num{56.58} to \num{59.65} under weak supervision, removing the remaining dependence on fully-supervised matching.
    \item We validate the approach on standard benchmarks, performing favorably compared to fully-supervised training methods.
\end{itemize}

\section{Related Work}
\label{sec:related_work}

\PAR{Hand-crafted keypoint extraction.}
Classical detectors and descriptors such as SIFT~\cite{Lowe.2004}, SURF~\cite{Bay.2006}, and ORB~\cite{Rublee.2011} rely on hand-crafted image statistics to identify repeatable interest points and encode their local appearance.
While highly efficient, these methods lack robustness under severe appearance changes such as day--night transitions or seasonal variation, motivating the shift toward learned alternatives.

\PAR{Supervised and self-supervised keypoint learning.}
The first wave of learned detectors coupled detection and description in a single end-to-end network.
SuperPoint~\cite{DeTone.2018} bootstraps training from synthetic shapes and refines via homographic adaptation,
D2-Net~\cite{Dusmanu.2019} jointly detects and describes dense feature maps,
R2D2~\cite{Revaud.2019} explicitly decouples repeatability and reliability,
and ALIKED~\cite{Wang.2023} introduces differentiable matching layers with attention-weighted local descriptors.
Li~\etal~\cite{Li.2022} demonstrated that decoupling detection and description mitigates the adverse influence of weak descriptors on detector accuracy, a paradigm further refined by DeDoDe~\cite{Edstedt.2024,Edstedt.2024hsa}, which learns detectors directly from 3D feature tracks extracted via SfM.
SiLK~\cite{Gleize.2023} and DomainFeat~\cite{Xu.2024f5b} additionally leverage style-transfer and triplet losses to bridge domain gaps induced by illumination changes.
A persistent limitation shared by all these methods is their dependence on MegaDepth~\cite{Li2018} or similar reconstruction-derived datasets, whose phototourism bias and reliance on SIFT-bootstrapped COLMAP~\cite{schoenberger2016sfm} models constrain generalization to challenging real-world conditions.

\PAR{Reinforcement learning for keypoint detection.}
The inherently discrete nature of keypoint selection prevents direct gradient-based optimization of the detector, motivating an RL formulation.
Bhowmik~\etal~\cite{Bhowmik.2020} first treated the downstream matching and pose estimation pipeline as a non-differentiable black box and computed rewards from known camera poses.
DISK~\cite{Tyszkiewicz.2020} refined this idea with per-match rewards derived from ground-truth depth, while DEAL~\cite{Potje.2023} extended DISK to handle non-rigid deformations.
S-TREK~\cite{101109iccv51070202300892} addressed patch-boundary artifacts via sequential off-policy sampling with equivariant convolutions, and DaD~\cite{dad2025} closed the resulting train-inference gap by aligning the sampling strategy across both stages, training a pure detector via policy-gradient repeatability objectives -- though still with depth-supervised reprojection rewards.
RaCo\cite{Shenoi2026} trains its detector from single-image homographies, requiring no multi-view supervision, yet still relies on the supervised ALIKED descriptor. This partial weak supervision contrasts with our approach, which trains both detector and descriptor from image pairs alone.
RIPE~\cite{knzel2025ripe-7fa} takes a different route:
it trains jointly on detection and description using only image pairs annotated with a same-scene/different-scene label, exploiting the epipolar constraint via RANSAC-based fundamental matrix estimation as a geometry-aware reward signal.
Our work builds directly on RIPE and addresses its two key limitations: the coarse binary reward and the reliance on negative training pairs.

\PAR{Feature matchers.}
Sparse keypoint correspondences must be established by a dedicated matcher, which itself requires training supervision.
SuperGlue~\cite{Sarlin.2020} introduced graph neural networks with self- and cross-attention to leverage geometric context from both images.
LightGlue~\cite{lindenberger_2023_lightglue} replaced this with a more efficient transformer architecture with adaptive early stopping.
Subsequent work explored state-space models~\cite{10.1109/icra55743.2025.11128473}, diffusion-based assignment~\cite{10.1145/3664647.3681069}, and foundation-model guidance~\cite{10.1109/cvpr52733.2024.01878,xuelun2024gim}.
While dense and semi-dense matchers such as RoMa~\cite{Edstedt.2024c99}, the DUSt3R family~\cite{Wang.2023s18,Leroy.2024}, and LoFTR~\cite{Sun.2021,Wang.2024} have pushed accuracy benchmarks, sparse matchers remain the dominant choice in production SfM and SLAM systems due to their favorable computational scaling, modularity, and compatibility with established pipelines.
Crucially, all existing sparse matchers are trained using pose or depth supervision, a dependency that mirrors and compounds the data constraints on the detector side.
In this work, we demonstrate that the weakly-supervised RL training paradigm can be extended to the matching step, removing this bottleneck.

\section{Method}
\label{sec:method}
In this section, we present our approach for weakly supervised keypoint detection and description from image pairs, without requiring geometric ground truth such as depth, camera pose, or annotated correspondences (\cref{subsec:extraction}). 
Our main contribution is a reformulation of the geometric reward signal used in reinforcement-learning-based keypoint learning, resulting in a simpler training process and improved results.
We additionally explore how the same learning scheme can be used to train a matcher (\cref{subsec:training_matcher}).

\subsection{Weakly-supervised Keypoint Extraction}
\label{subsec:extraction}

We begin with a brief recap of RIPE~\cite{knzel2025ripe-7fa} to establish the foundation for our modifications. We then show how to train a keypoint extractor using paired images only, to reshape the reward to reflect actual match quality and how to enforce precise keypoint localization in the heatmap through entropy regularization.

\subsubsection{Recap on RIPE}

\begin{figure}[tb]
    \centering
    \includegraphics[width=1.0\linewidth]{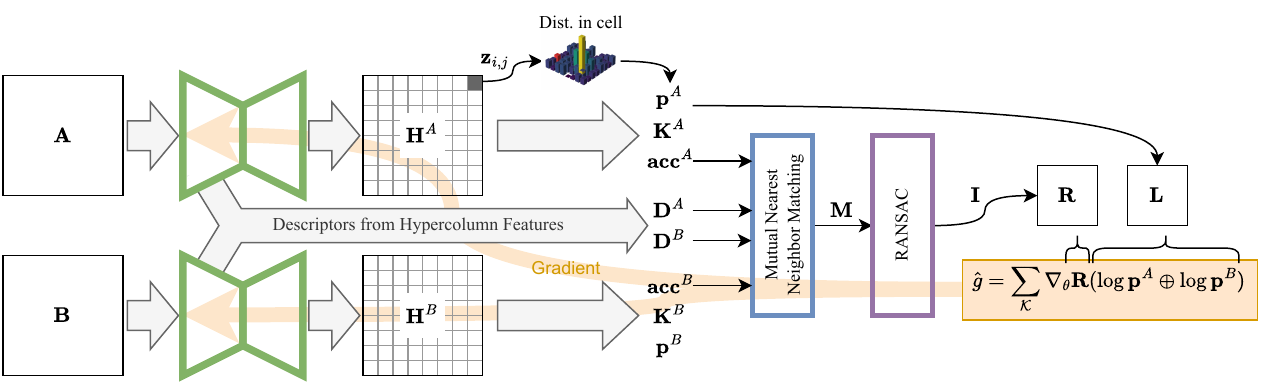}
    \caption{Schematic overview of the proposed weakly-supervised keypoint extraction framework. During training, the framework operates on pairs of images, from which a neural network generates heatmaps $\mathbf{H}$ from which keypoint locations $\mathbf{K}$ with a probability $\mathbf{p}$ are sampled. Descriptors $\mathbf{D}$ are extracted from the encoder, as Hypercolumn Features. Using these descriptors, keypoints are matched and filtered using epipolar constraints to calculate the reward matrix $\mathbf{R}$, which weights the combined log-probabilities $\mathbf{L}$ of the keypoint locations, resulting in an approximation of the gradients, using REINFORCE~\cite{williams.1992}.}
    \label{fig:schema_ripe}
\end{figure}

RIPE~\cite{knzel2025ripe-7fa} demonstrates how RL can be used to train a keypoint extractor using only image pairs that are labeled as either positive (showing the same scene, \ie 2D projections of 3D points exist in both images) or negative (images are from different scenes).
An overview is shown in \cref{fig:schema_ripe}.
Given an input image $\mathbf{A} \in \mathbb{R}^{h \times w \times c}$, a neural network predicts a heatmap $\mathbf{H}^A \in \mathbb{R}^{h \times w}$ indicating potential keypoint locations.
During training, the heatmap $\mathbf{H}^A$ is divided into $n$ quadratic cells of size $q$.
The logit values, within a cell, define a categorical distribution, from which exactly on keypoint location is sampled.
These sampled keypoint locations $\mathbf{K}^A \in \mathbb{R}_{2 \times n}$ are associated with initial probabilities $\mathbf{\hat{p}} \in \mathbb{R}_{1 \times n}$ (calculated from the distribution) and corresponding logit values $\mathbf{z}^A \in \mathbb{R}_{1 \times n}$.
This formulation allows the network to shape the heatmap distribution according to the reward signal.
To account for unreliable locations (sky, texture-less or over-exposed regions), RIPE introduces an acceptance indicator $\textbf{acc}^A=\text{Sigmoid}(\mathbf{z}^A)$ to reject keypoint locations based on their logit value.
The final probability of a keypoint is calculated as $\mathbf{p}^A=\mathbf{\hat{p}^A} \odot \textbf{acc}^A$, with $\odot$ denoting the element-wise product.
Keypoint locations are then associated with their descriptors $\mathbf{D}^A$, computed as hypercolumn descriptor~\cite{Hariharan.2015} from the layers of the encoder part of the network (see~\cite{knzel2025ripe-7fa} for details).
During training, the same procedure yields acceptance indicators $\textbf{acc}^B$, location $\mathbf{K}^B$ and probabilities $\mathbf{p}^B$, for a paired image $\mathbf{B}$.
Using REINFORCE~\cite{williams.1992}, the gradient for maximizing the expected reward can be approximated as

\begin{equation}
    \nabla_\theta \mathbb{E}_\mathcal{K}[\mathbf{R}] \approx \hat{g} = \sum_{\mathcal{K}} \nabla_\theta \mathbf{R} (\log \mathbf{p}^A \oplus \log \mathbf{p}^B) ,
\label{eq:reinforce}
\end{equation}

\noindent with $\oplus$ denoting the outer sum, resulting in the combined matrix of log-probabilities $\mathbf{L}\in \mathbb{R}^{n \times n}$ and $\mathcal{K}$ representing all possible combinations of keypoints.
Note, that this formulation could also be viewed as a Contextual Bandit~\cite{Lattimore_Szepesvari_2020}, as each image pair constitutes a context, on which the drawn actions (keypoint locations) directly depend and receive an immediate reward.

\begin{figure}[tb]
    \centering
    \includegraphics[width=1.0\linewidth]{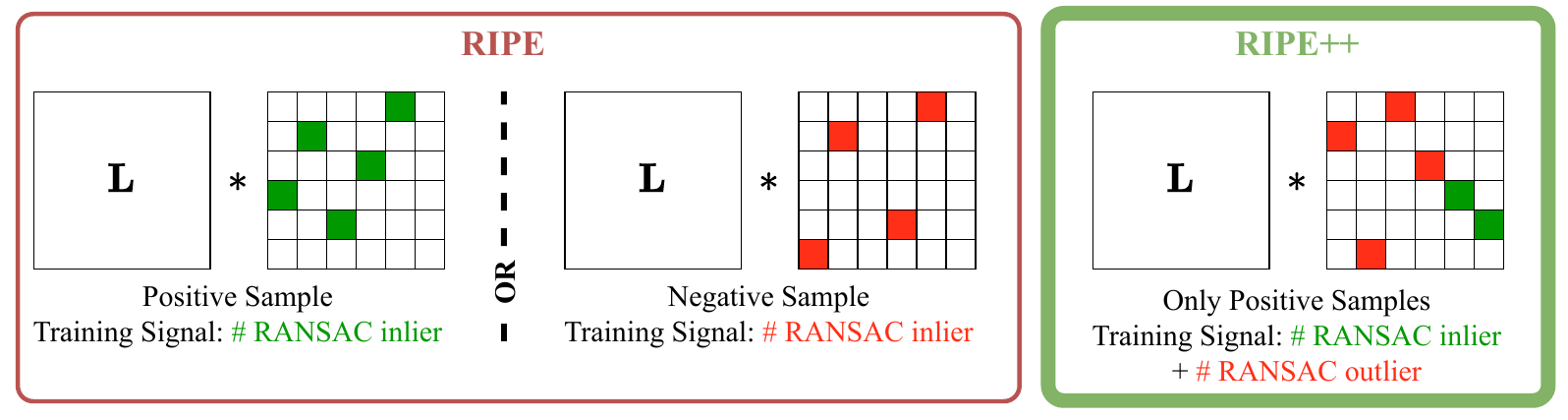}
    \caption{Visual comparison of the reward matrix construction to shape the log-probabilities in $\mathbf{L}$ for the gradient computation (\cref{eq:reinforce}). In RIPE (left) rewards are assigned at image pair level: geometric consistent matches are rewarded for positive pairs and penalized for negative sample pairs, outliers are ignored. In RIPE\texttt{++} (right), rewards are defined at correspondence level within positive pairs only. Geometrically consistent matches are rewarded and geometrically inconsistent matches are explicitly penalized. This finer-grained reward makes negative pairs unnecessary and yields a more informative training signal.}
    \label{fig:reward_matrices}
\end{figure}

RIPE`s key component lies in the calculation of the reward matrix $\mathbf{R} \in \mathbb{R}^{n \times n}$, which translates geometric consistency into a learning signal.
The reward is derived from the number of detected keypoints that can be matched and are geometrically plausible.
Concretely, mutual nearest-neighbors are established between descriptors, yielding a list $\mathbf{M} \in \mathbb{R}^{m \times 2}$ of index pairs for matched keypoints, with $m<<n$. 
Subsequently, these candidates are filtered by a robust estimation of the fundamental matrix (assuming rigid transformations only and pinhole cameras), resulting in a list of inlier indicators $\mathbf{I} \in \mathbb{R}^{m}$.
The reward matrix $\mathbf{R}$, is then constructed by

\begin{equation}
r_{p,q} = 
\begin{dcases} 
\rho_{\text{in}}, & (i, j) \in \mathbf{M} \text{ and } \mathbf{I}_{i,j} \text{ is True} \\ 
\lambda, & \text{otherwise}
\end{dcases}
\label{eq:reward}
\end{equation}

\noindent with $\rho$ being the reward and $\lambda$ being a small negative penalty for not finding a sufficient keypoint.
For a negative image pair, the reward is inverted to penalize the network for finding geometrically plausible matches ($\rho_{\text{out}}=-\rho_{\text{in}}$) for a negative pair, and give a small reward (positive $\lambda$) for detecting only keypoints not resulting in plausible matches.
A visualization of the resulting reward matrices is shown in \cref{fig:reward_matrices} (left).
As a result, the gradients from \cref{eq:reinforce} allow the network to shape the heatmap in such a way that geometrically consistent correspondences in positive pairs have a higher probability of being selected and vice versa for the negative case.

\subsubsection{Training From Positive Samples Only}

The reliance on both positive and negative pairs in RIPE is closely tied to the structure of its reward formulation. Since the reward for positive pairs is solely based on counting geometrically consistent inliers, negative pairs are required to introduce a counter-signal that penalizes such matches.
This design, however, has several limitations.
First, mislabeled negative pairs -- images incorrectly marked as depicting different scenes -- can mislead the detector and can destabilize training.
Second, and more fundamentally, the inversion of the reward for negative pairs penalizes only the number of \textit{geometrically consistent} matches.
As a consequence, the network is able to produce many false matches without penalty, as long as they are filtered out by RANSAC.
In the extreme case where all keypoints are matched but subsequently rejected by robust estimation, the network receives a neutral gradient signal, resulting in a degraded training performance.
We therefore reformulate the training objective to use only positive pairs, directly rewarding inlier matches and penalizing outliers.

This requires revisiting the reward formulation itself.
Instead of counting inliers, we fully exploit the supervision from the robust fundamental matrix estimation and assign a reward $\rho_{\text{in}}$ to every element $r_{i, j}$ in $\mathbf{R}$ that was identified to be an inlier and a penalty $\rho_{\text{out}}$ to outliers. 
Extending \cref{eq:reward} gives

\begin{equation}
r_{i, j} = 
\begin{dcases} 
\rho_{\text{in}}, & (i, j) \in \mathbf{M} \text{ and } \mathbf{I}_{i, j} \text{ is True} \\
\rho_{\text{out}}, & (i, j) \in \mathbf{M} \text{ and } \mathbf{I}_{i, j} \text{ is False} \\
\lambda, & \text{otherwise}
\end{dcases},
\label{eq:reward_pos}
\end{equation}

\noindent for learning from positive samples only.
Beyond simplifying supervision, this reformulated reward also enables the matcher training in \cref{subsec:training_matcher}.

\subsubsection{Entropy Regularization}

To prevent the generation of low-probability keypoints, RIPE introduces, inspired by Potje~\etal~\cite{Potje.2023}, a regularization term $\mathcal{L}_{\text{low}}=-\sum_i \log p_{i} \cdot \epsilon,$ where $\epsilon$ is a small negative constant and $p_i \in \mathbf{p}$ denotes the logit values in the $i$-th cell.
While this regularization has been proven effective, we observe that the resulting heatmaps lack sufficient spatial precision, particularly at lower input resolutions (see \cref{subsec:ablations}).
Although $\mathcal{L}_{\text{low}}$ encourages high probability for the selected keypoint, it only implicitly shapes the surrounding probability landscape: since keypoints are sampled from a categorical distribution over each patch, a high probability for the selected location necessitates low probabilities elsewhere, but does not explicitly enforce a concentrated distribution.
To address this, we replace $\mathcal{L}_{\text{low}}$ with the negative entropy
\begin{equation}
    \mathcal{L}_{\text{H}}=- \sum_i \sum_j \mathbf{p}_{i,j} \log \mathbf{p}_{i,j},
    \label{eq:entropy_regularization}
\end{equation}
\noindent where $\mathbf{p}_{i,j}$ represents the probabilities of the individual keypoint locations in one cell.
This directly encourages the distribution within each patch to approach a one-hot encoding.
Unlike $\mathcal{L}_{\text{low}}$, minimizing $\mathcal{L}_{\text{H}}$ produces gradients for all positions in the patch, explicitly shaping the entire distribution rather than only penalizing the selected keypoint.

\subsubsection{Final Loss}

We define the loss of the detector as $\mathcal{L}_{\text{dect}}=-\mathbb{E}_{\mathcal{K}}[\mathbf{R}]$.
As the Reinforcement Learning only optimizes the locations of the keypoints, but not their descriptors, we utilize the contrastive descriptor loss $\mathcal{L}_{\text{desc}}$ from RIPE\cite{knzel2025ripe-7fa} to pull the descriptors of putative matches closer and repel others.
Combining these with our introduced entropy regularization results in 
$\mathcal{L} = \mathcal{L}_{\text{dect}} + \omega \mathcal{L}_{H} + \psi \mathcal{L}_{\text{desc}}$, with $\omega$ and $\psi$ balancing the influence of the regularization and the descriptor loss respectively.

\subsubsection{Further Modifications}
Beyond the core contributions described above, we explored several additional directions, including curriculum learning strategies and alternative descriptor loss formulations.
We document these investigations in the supplementary material, as they provide useful insights for practitioners working on RL-based keypoint learning.

\subsection{Extension to Weakly-supervised Matcher}
\label{subsec:training_matcher}

\begin{figure}[tb]
    \centering
    \includegraphics[width=1.0\linewidth]{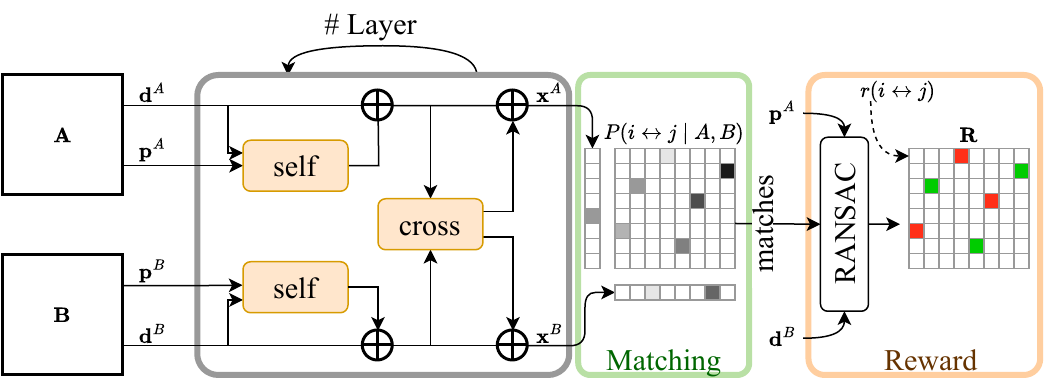}
    \caption{Schematic overview of the proposed weakly-supervised training framework for LightGlue. Given keypoint locations and descriptors from two images, LightGlue refines the representations through multiple layers of self- and cross-attention, incorporating positional and cross-image context. Rather than supervising with ground-truth correspondences derived from depth or pose data, we formulate matching as a reinforcement learning problem: matches deemed geometrically consistent by RANSAC receive a positive reward, while outliers are penalized. This enables end-to-end training of the matcher using only image pairs.}
    \label{fig:schema_lightglue}
\end{figure}

To be competitive, modern sparse keypoint extractors rely on dedicated, trained matchers such as SuperGlue~\cite{DeTone.2018} or
LightGlue~\cite{lindenberger_2023_lightglue}.
However, these methods require pose or depth annotations for training, which contradicts our goal of learning from weak supervision via image pairs alone.
As an additional experiment, we therefore explore whether the same geometric reward can also be used to train a transformer-based matcher. Concretely, we adapt LightGlue to enable training using only geometric consistency as a reward signal, drawing on the policy gradient formulation of DISK~\cite{Tyszkiewicz.2020}.
\cref{fig:schema_lightglue} illustrates our approach.

\subsubsection{Recap on LightGlue}

Given two sets of local features from images $A$ and $B$, each with normalized 2D positions $\mathbf{p}_i$ and descriptors $\mathbf{d}_i \in \mathbb{R}^d$, LightGlue refines point representations through $L$ stacks of self- and cross-attention layers.
A lightweight head then computes pairwise scores
\begin{equation}
    S_{i,j} = \text{Linear}\left(\mathbf{x}_i^A\right)^\top
    \text{Linear}\left(\mathbf{x}_j^B\right),
    \label{eq:sim_lightglue}
\end{equation}
\noindent and per-point matchability scores $\sigma_i = \text{Sigmoid}\left(\text{Linear}\left(\mathbf{x}_i\right)\right) \in [0,1]$, which are combined into a soft partial assignment matrix
\begin{equation}
    P_{i,j} = \sigma_i^{A} \, \sigma_j^{B} \,
    \underset{k \in \mathcal{A}}{\text{Softmax}}(\mathbf{S}_{k,j})_i \;
    \underset{k \in \mathcal{B}}{\text{Softmax}}(\mathbf{S}_{i,k})_j.
    \label{eq:partial_assignment_lightglue}
\end{equation}

\subsubsection{Weakly-supervised  LightGlue}
The original LightGlue loss maximizes the log-likelihood of ground-truth assignments, requiring known correspondences from pose or depth.
To avoid this supervision and allow training from image pairs only, we reformulate the optimization using a reward-driven objective.
Inspired by DISK~\cite{Tyszkiewicz.2020}, we develop a policy gradient formulation, interpreting matching as a decision process and optimize the expected geometric reward instead of a supervised assignment loss.
Concretely, we decompose \cref{eq:partial_assignment_lightglue} into the probability of a match,

\begin{equation}
    P(i,j \mid A, B) =
    \underset{k \in \mathcal{A}}{\text{Softmax}}(\mathbf{S}_{k,j})_i \;
    \underset{k \in \mathcal{B}}{\text{Softmax}}(\mathbf{S}_{i,k})_j,
    \label{eq:prob_match_matcher}
\end{equation}
and 
\begin{equation}
    P_i(\mathbf{x}_i \mid I) = \text{Sigmoid}\left(\text{Linear}\left(\mathbf{x}_i\right)\right),
    \label{eq:not_matchable}
\end{equation}

\noindent the probability of accepting keypoint $i$ as matchable.
The expected reward gradient then takes the form
\begin{equation}
    \nabla_\theta \, \underset{A \leftrightarrow B}{\mathbb{E}} \,
    R(M_{A \leftrightarrow B}) =
    \underset{F_A, F_B}{\mathbb{E}} \sum_{i,j}
    P(i,j \mid A, B) \cdot r(i,j)
    \cdot \nabla_{\theta} \Gamma_{ij},
    \label{eq:ripe_match_gradient}
\end{equation}
where $r(i \leftrightarrow j)$ is a per-match reward based on geometric consistency and $\Gamma_{ij} = \log P(i \leftrightarrow j \mid A, B) + \log P_i(\mathbf{x}_i \mid A) + \log P_j(\mathbf{x}_j \mid B)$.
Since all match probabilities can be computed in closed form, no sampling is required and the expectation is exact up to the empirical approximation over feature sets $F_A, F_B$. 
We provide a detailed derivation of the connection between the DISK and LightGlue formulations in the supplementary material.

The reward signal $r(i,j)$ is  derived from the geometric consistency of the predicted correspondences. To this end, we estimate the fundamental matrix using a robust estimator and classify the resulting correspondences as inliers or outliers. Inliers receive a reward $\nu_{\text{in}}$, while outliers receive a penalty $\nu_{\text{out}}$.
Since this reward signal only applies to matched keypoints, the network could trivially minimize the loss by predicting all keypoints as non-matchable. 
To avoid this degenerate solution, we add a small penalty $\lambda$ for each keypoint classified as unmatchable.

To prevent the network from trivially labeling every keypoint as not matchable according to  \cref{eq:not_matchable} we introduce a regularization term $\mathcal{L}_{\text{nm}} = \sum_i P_i(\mathbf{x}_i|A) + \sum_i P_i(\mathbf{x}_i|B)$ which corresponds to the expected number of keypoints the model classifies as non-matchable.
We define the loss of the matcher as $\mathcal{L}_{\text{match}}=-\mathbb{E}_{M_{A\leftrightarrow B}}[R(M_{A\leftrightarrow B})]$ and the final training loss as

\begin{equation}
    \mathcal{L}=\mathcal{L}_{\text{match}} + \eta \mathcal{L}_{\text{nm}},
    \label{eq:final_loss_matcher}
\end{equation}

\noindent with $\eta$ controlling the strength of the non-matchable regularization term.

\section{Experiments}
\label{sec:experiments}

\PAR{Implementation Details}
To ensure a fair comparison with RIPE, we use the same \textsc{torchvision} VGG-19 backbone combined with the depthwise convolutional refiners proposed by Edstedt~\etal~\cite{10.48550/arxiv.2202.00667}.
We use the GPU-accelerated implementation of nearest-neighbor matching from Kornia~\cite{eriba2019kornia}.
For robust fundamental matrix estimation, we replace PoseLib~\cite{PoseLib} with the \textsc{usac\_magsac} RANSAC variant from OpenCV~\cite{opencv_library}, which achieves comparable accuracy at lower computational cost, reducing overall training time.
The keypoint extractor is trained with AdamW~\cite{Loshchilov2017DecoupledWD} using a learning rate that decays linearly from $10^{-3}$ to $10^{-6}$ over 80k steps.
We use  a batch size of 6 with gradient accumulation over 4 batches, and no data augmentation beyond normalization, resizing the longer side to 560 pixels and padding to preserve the aspect ratio.
As usual, the cell size is set to 8 pixels.
Training takes 26 hours on a single A100, compared to 72 hours for RIPE -- another benefit of using positive samples only, since RANSAC can exit early for positive samples but must run the maximum iterations for negative ones.
Please refer to the supplementary for the remaining hyperparameters.

\PAR{Training data}
To enable a controlled comparison with existing methods and to isolate the effects of our training scheme, we train on the MegaDepth dataset, using the subset proposed by DISK~\cite{Tyszkiewicz.2020} for the keypoint extractor and the subset proposed by LightGlue~\cite{lindenberger_2023_lightglue} for the matcher.
Since our method requires only image pairs without depth or pose annotations, it naturally generalizes to domains where such supervision is notoriously difficult to acquire; we demonstrate this by training on endoscopic video streams in \cref{subsec:scared1500}.

\PAR{Inference}
While training operates on image pairs, at inference the keypoint extractor processes a single image: the top-$k$ keypoints are selected from the heatmap $\mathbf{H}$ based on their score, after applying non-maximum suppression in a $3\times3$ window and subpixel refinement following ALIKED~\cite{Zhao.2023dvq}, as also adopted by RaCo~\cite{Shenoi2026} and DaD~\cite{dad2025}.

\subsection{Relative Pose Estimation}
\label{subsec:rel_pose_eval}

\PAR{Dataset} We evaluate relative pose estimation on the MegaDepth-1500 benchmark introduced by LoFTR~\cite{Sun.2021}, which comprises two scenes (\textit{Brandenburger Tor} and \textit{St.\ Peters Square}) from the 196-scene MegaDepth dataset.
The benchmark poses challenges through large viewpoint changes, varying illumination, and repetitive structures.
We resize the longer image side to 1200 pixels and to isolate the quality of the extracted keypoints, all sparse methods are evaluated using mutual-nearest-neighbor matching.

\PAR{Metrics}
Following standard practice, match quality is measured indirectly through the accuracy of the resulting relative poses.
The pose error is defined as the maximum of the angular errors in rotation and translation.
We report the Area Under the Curve (AUC) of the pose error at thresholds of \ang{5}, \ang{10}, and \ang{20}.
All experiments use the top 2048 keypoints and robust pose estimation via PoseLib~\cite{PoseLib}, implemented within the glue-factory framework~\cite{pautrat_suarez_2023_gluestick,lindenberger_2023_lightglue}.

\PAR{Baselines}
We compare RIPE\texttt{++} against SotA methods for sparse keypoint extraction.
As RaCo and DaD are only keypoint detectors, we combine both with ALIKED-n16 descriptors for evaluation.
For the all methods, we use mutual-nearest-neighbor matching to establish correspondences, and a RANSAC inlier threshold of \num{0.5}.

\begin{table}[tb]
\centering
\setlength{\aboverulesep}{0pt}
\setlength{\belowrulesep}{0pt}
\setlength{\extrarowheight}{0.75ex}
\scriptsize
\caption{Relative pose estimation on MegaDepth1500. Methods are grouped by whether training requires geometric ground truth (pose, depth) for the detector or descriptor. RIPE\texttt{++} trains both detector and descriptor from positive image pairs alone, without geometric supervision, yet remains comparable to geometric ground truth group and outperforms RIPE despite discarding its negative pairs. \textbf{Best} and \underline{second-best} are highlighted within each group.}
\begin{tabular*}{\textwidth}{@{\extracolsep{\fill}}c l
    S[table-format=2.2]
    S[table-format=2.2]
    S[table-format=2.2]
    S[table-format=2.2]
    @{}}
\toprule
& Method & {AUC$@5^{\circ}$} & {AUC$@10^{\circ}$} & {AUC$@20^{\circ}$} & {Supervision} \\
\midrule
\multirow{5}{*}{\rotatebox[origin=c]{90}{Geometric GT}}
& ALIKED\cite{Zhao.2023dvq}\tiny{TIM'23} & {\underline{56.66}} & 69.64 & 79.35 & {Pose+Homog.}\\
& DaD\cite{dad2025}\tiny{CoRR'25} & 56.46 & {\underline{70.07}} & {\underline{80.18}} & {Depth}\\
& DeDoDe-B\cite{Edstedt.2024hsa}\tiny{CVPRW'24} & 56.01 & 69.20 & 78.72 & {Pose}\\
& DISK\cite{Tyszkiewicz.2020}\tiny{NeurIPS'20} & 50.69 & 64.35 & 74.74 & {Pose/ Depth}\\
& RACO\cite{Shenoi2026}\tiny{3DV'26} & {\bfseries 57.96} & {\bfseries 71.28} & {\bfseries 81.00} & {Homog. (Det.) + Pose (Desc.)}\\
\midrule
\multirow{4}{*}{\rotatebox[origin=c]{90}{No geo. GT}}
& SuperPoint\cite{DeTone.2018}\tiny{CVPRW'18} & 47.26 & 60.89 & 70.65 & {Homography}\\
& SIFT\cite{Lowe.2004}\tiny{IJCV'04} & 38.97 & 53.21 & 64.99 & {---}\\
& RIPE\cite{knzel2025ripe-7fa}\tiny{ICCV'25} & {\underline{53.47}} & {\underline{67.02}} & {\underline{77.62}} & {Pos. + Neg. Pairs}\\
& {\bfseries RIPE++} & {\bfseries 56.58} & {\bfseries 69.53} & {\bfseries 79.33} & {Pos. Pairs}\\
\bottomrule
\end{tabular*}
\label{tab:megadepth}
\end{table}

\begin{figure}[tb]
    \centering
    \begin{subfigure}[b]{\textwidth}
            \includegraphics[height=1.57cm]{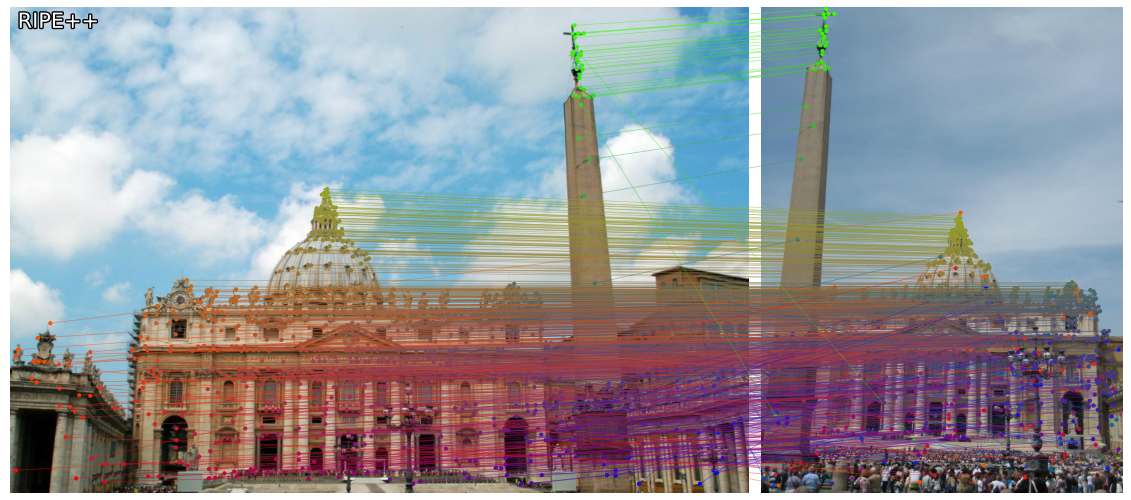}
            \includegraphics[height=1.57cm]{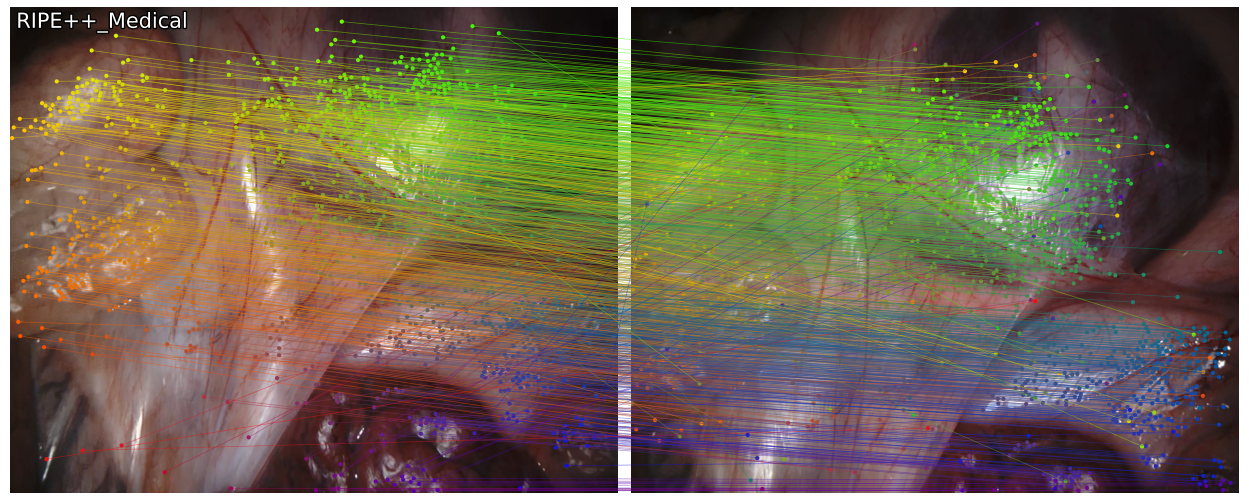}
            \includegraphics[height=1.57cm]{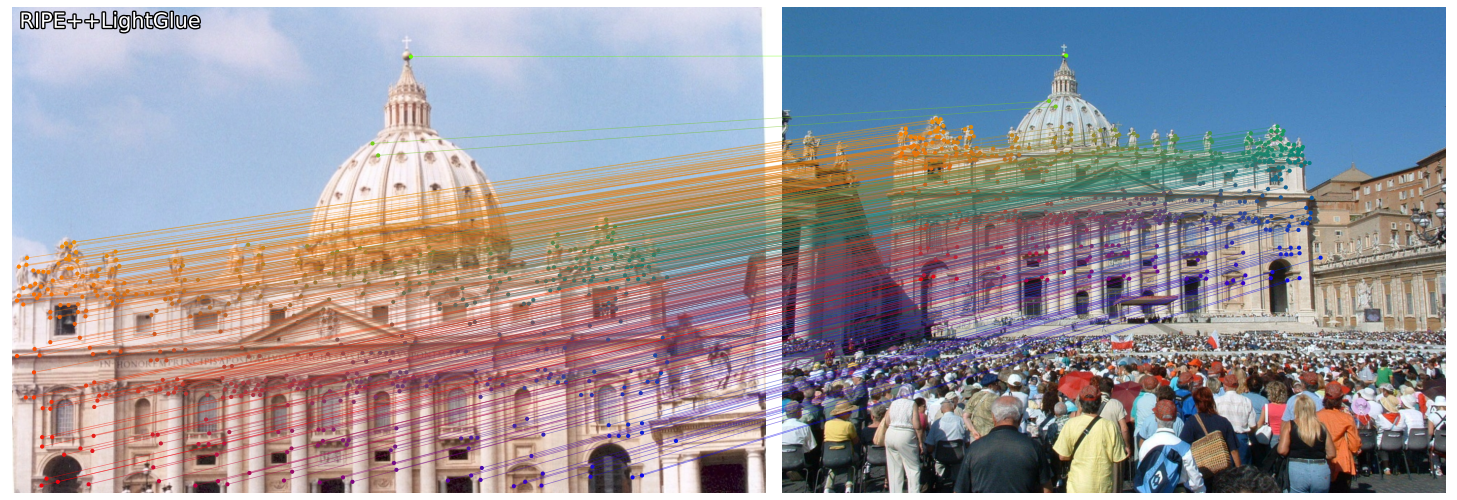}
    \end{subfigure}
\caption{Qualitative results of raw matches for RIPE\texttt{++} and RIPE\texttt{++} in combination with our weakly-supervised LightGlue matcher on the MegaDepth1500 benchmark dataset on the left and right respectively. The middle image shows qualitative results for RIPE\texttt{++}~Medical on the SCARED1500 test dataset. For more qualitative results please refer to the supplementary material.}
\label{fig:qualitative_results_main}
\end{figure}

\PAR{Results}
Tab.~\ref{tab:megadepth} groups methods by whether their training requires geometric ground truth. Among methods that use none, RIPE\texttt{++} is the strongest by a clear margin, improving over RIPE by \SI{3.11}{pp} AUC@5° while discarding the negative pairs RIPE depends on. More notably, this parity extends across the supervision boundary: RIPE\texttt{++} matches the strongly-supervised extractors, trailing ALIKED by only \SI{0.07}{pp} on average and outperforming DeDoDe\cite{Edstedt.2024hsa} and DaD\cite{dad2025}, both of which use separate, fully-supervised detection and description networks. RaCo\cite{Shenoi2026} reports higher absolute numbers, but only its detector is trained without pose supervision; its descriptor is taken from the supervised ALIKED. RIPE\texttt{++}, by contrast, trains both detector and descriptor from image pairs alone, without artificial homographies, pose annotations, or depth from pre-registered 3D models.
Qualitative results are provided in \cref{fig:qualitative_results_main} and the supplementary material.

\subsection{SCARED1500}
\label{subsec:scared1500}
The weak supervision of RIPE\texttt{++} enables training keypoint extractors in domains where ground-truth pose or depth information is difficult to obtain, such as for instance endoscopic imaging.
We demonstrate this by training directly from endoscopic camera streams, which are readily available in large quantities.

\begin{table}[tb]
\centering
\setlength{\tabcolsep}{4pt}
\setlength{\aboverulesep}{0pt}
\setlength{\belowrulesep}{0pt}
\setlength{\extrarowheight}{0.75ex}
\caption{Evaluation of the relative pose estimation on Scared 1500, showing the benefits of our proposed training scheme. It allows to easily train a specialized keypoint extractor, simply from video frames. The \textbf{best} and \underline{second-best} performances are highlighted.}
\scriptsize
\begin{tabular*}{\textwidth}{@{\extracolsep{\fill}}l
    S[table-format=2.2]
    S[table-format=2.2]
    S[table-format=2.2]
    S[table-format=2.2]}
\toprule
Method & {AUC$@5^{\circ}$} & {AUC$@10^{\circ}$} & {AUC$@20^{\circ}$} & {Supervision} \\
\midrule
ALIKED\cite{Zhao.2023dvq}\tiny{TIM'23} & 18.24 & 40.61 & 61.23 & {Pose+Homography}\\
DaD\cite{dad2025}\tiny{CoRR'25} & 18.44 & 40.56 & 61.40 & {Depth}\\
DeDoDe-B\cite{Edstedt.2024hsa}\tiny{CVPRW'24} & 18.85 & 41.69 & 62.80& {Pose} \\
DISK\cite{Tyszkiewicz.2020}\tiny{NeurIPS'20} & 16.58 & 36.77 & 55.63 & {Pose/ Depth}\\
RACO\cite{Shenoi2026}\tiny{3DV'26} & {\underline{19.35}} & 42.36 & 63.34 & {Homography}\\
SIFT\cite{Lowe.2004}\tiny{IJCV'04} & 13.24 & 32.17 & 51.66 & {---}\\
SuperPoint\cite{DeTone.2018}\tiny{CVPRW'18} & 19.01 & {\underline{42.71}} & {\underline{64.54}} & {Homography} \\
RIPE\cite{knzel2025ripe-7fa}\tiny{ICCV'25} & 17.43 & 38.63 & 58.70 & {Pos. + Neg. Pairs}\\
\midrule
RIPE\texttt{++} & 17.53 & 38.09 & 57.86 & {Pos. Pairs}\\ 
RIPE\texttt{++}~Medical & {\bfseries 20.90} & {\bfseries 46.51} & {\bfseries 68.72} & {\bfseries Video Data} \\
\bottomrule
\end{tabular*}
\label{tab:scared1500}
\end{table}

\begin{table}[tb]
\caption{Ablation study of our proposed methods on the MegaDepth1500 test set.}
\scriptsize
\begin{tabularx}{\textwidth}{CCCCCCC}
\toprule
\multicolumn{2}{c}{\textbf{Parameter}} & \multicolumn{5}{c}{\textbf{Metrics}} \\ 
\cmidrule(lr{.5em}){1-2} \cmidrule(lr{.5em}){3-7}
{pos only} & {entropy} & {\# RANSAC inlier} & { \% RANSAC inlier} & {AUC$@5^{\circ}$} & {AUC$@10^{\circ}$} & {AUC$@20^{\circ}$} \\
\cmidrule(lr{.5em}){1-2} \cmidrule(lr{.5em}){3-7}
               &  --           & 297.5        & 34.4          & 51.83                & 65.37                 & 75.94                 \\
{\cmark}       &  --           & 427.5        & 41.5          & 52.42                & 66.78                 & 78.03                 \\
{\cmark}       & \num{1e-4}    & {\text{--}}  & {\text{--}}   & {\text{--}}          & {\text{--}}           & {\text{--}}           \\
{\cmark}       & \num{1e-5}    & 266          & 33.6          & 49.62                & 62.7                  & 72.8                  \\
\cmidrule(lr{.5em}){1-2} \cmidrule(lr{.5em}){3-7}
{\cmark}       & \num{1e-6}    & 366.5        & 40.6          & 56.59                & 69.44                 & 79.18                 \\ \bottomrule
\end{tabularx}
\label{tab:ablations}
\end{table}

\PAR{Dataset}
We introduce SCARED1500, a benchmark derived from the SCARED dataset~\cite{scared2019}, recorded with a da Vinci Xi surgical robot with accurate camera poses, which we only use for our evaluation.
The dataset contains 7 training and 2 test scenes of porcine subjects, each with 4--5 keyframes representing unique viewpoints with structured-light depth.
Test pairs are formed by randomly sampling image pairs within the same keyframe sequence and retaining only those with 40--90\% visual overlap, estimated by projecting 3D points between frames.
Images are undistorted using the provided calibration and resized to 1200 pixels on the longer side at inference.
Training pairs for RIPE\texttt{++} are formed by frames separated by 60 frames, yielding 17\,514 pairs, augmented with random affine transformations (in-plane rot. within ±30°, transl. up to \SI{10}{\percent} of image size, and isotropic scaling in [0.9,1.1]) to mitigate overfitting.
While this simple pairing strategy can produce degenerate pairs with negligible motion or no overlap, we find the model robust to such noise given a reasonable frame distance.

\PAR{Metrics}
We integrate SCARED1500 into the glue-factory framework to leverage its evaluation pipeline.
As in \cref{subsec:rel_pose_eval}, we evaluate relative pose estimation using the same metrics, with a RANSAC threshold of 1.0.

\begin{figure}[tb]
    \centering
    \includegraphics[width=0.8\linewidth]{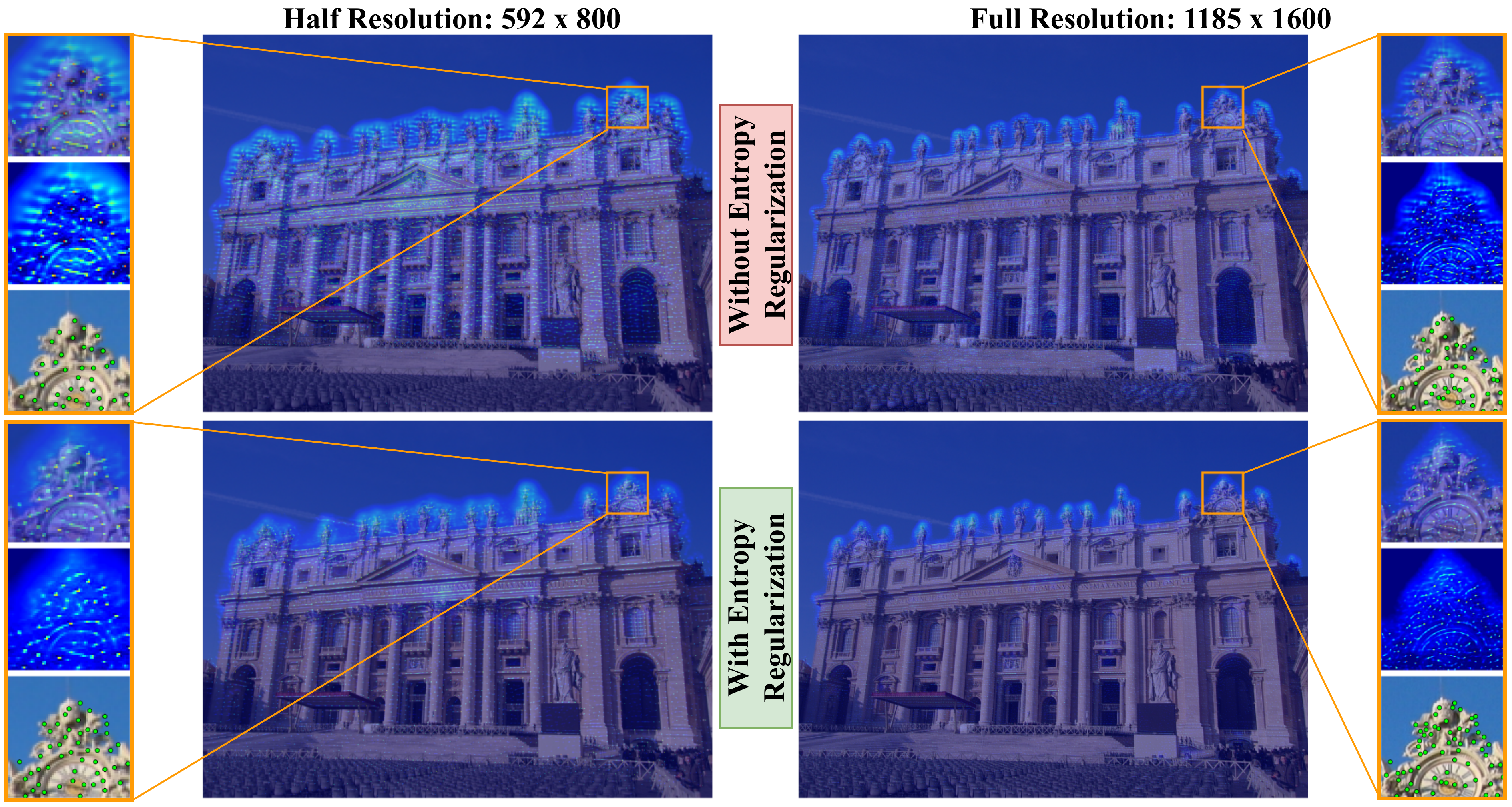}
    \caption{Effect of our entropy-based regularization on keypoint localization. Without it, heatmaps tend to become diffuse, particularly at low input resolutions (top-left). Regularization encourages sharper, more peaked responses (bottom).}
    \label{fig:heatmap_entropy}
\end{figure}

\PAR{Results}
The value of weak supervision is clearest under domain shift (Tab.~\ref{tab:scared1500}). Retrained on endoscopic video, RIPE\texttt{++}~Medical outperforms all baselines across every threshold, adapting both detector and descriptor from raw frames without pose or depth. This adaptability is the operative advantage: zero-shot, every learned method collapses on the unseen domain, and RIPE\texttt{++} itself only matches RIPE, since its positive-only reward specializes to the training distribution rather than generalizing across domains. What distinguishes RIPE\texttt{++} is not zero-shot transfer but the ability to retrain cheaply wherever geometric ground truth is unavailable, which no pose- or depth-supervised method can do here. RaCo\cite{Shenoi2026} could in principle fine-tune its homography-trained detector, but its supervised ALIKED descriptor cannot follow, leaving the adaptation incomplete.
Qualitative results are provided in \cref{fig:qualitative_results_main} and the supplementary material.

\subsection{RIPE\texttt{++} with Learned Matcher}

To demonstrate the versatility of our RL-based training approach, we trained LightGlue as dedicated matcher on top of our learned RIPE\texttt{++} extractor, by extending the official LightGlue~\cite{lindenberger_2023_lightglue} implementation. We integrated our proposed training method, but retained the original two-stage training protocol: pretraining on synthetic image pairs generated from the distractor dataset~\cite{RITAC18}, followed by fine-tuning on MegaDepth using the official data split.
Since the inlier-to-outlier ratio during matcher training is more balanced than during extraction, we use rewards and penalties of equal magnitude.
\cref{tab:lightglue_matcher} shows the clear improvements gained for the MegaDepth1500 test set.
While the absolute pose accuracy remains below that of LightGlue variants trained with full pose and depth supervision (\eg, 66.1\% AUC$@5^{\circ}$ when paired with ALIKED), our matcher is trained entirely from image pairs without any geometric ground truth, making the average improvement 3.9\,pp a promising indication that RL-based weak supervision can transfer effectively to the matching stage.

\begin{table}[tb]
\centering
\setlength{\aboverulesep}{0pt}
\setlength{\belowrulesep}{0pt}
\setlength{\extrarowheight}{0.75ex}
\scriptsize
\caption{Improvements on MegaDepth1500 from training LightGlue with our proposed approach on positive image pairs only.}
\begin{tabularx}{\textwidth}{>{\raggedright\arraybackslash}X
    >{\centering\arraybackslash}X
    >{\centering\arraybackslash}X
    >{\centering\arraybackslash}X}
\toprule
Method & {AUC$@5^{\circ}$} & {AUC$@10^{\circ}$} & {AUC$@20^{\circ}$} \\
\midrule
RIPE\texttt{++} & 56.58 & 69.53 & 79.33 \\
$\downarrow$\scriptsize{ + LightGlue} & \gss{+3.07} & \gss{+4.16} & \gss{+4.53} \\
Matched-RIPE\texttt{++}& 59.65 & 73.69 & 83.86 \\
\bottomrule
\label{tab:lightglue_matcher}
\end{tabularx}
\end{table}

\subsection{Ablations}
\label{subsec:ablations}

To ablate the influence of our proposed methods, we evaluated our different configurations on the MegaDepth1500 test set with the configurations described in \cref{subsec:rel_pose_eval}.
\Cref{tab:ablations} shows that our improved reward calculation purely based on positive pairs not only simplifies the data generation, but also improves over training from positive and negative pairs.
Adding our regularization to ensure a precise localization of keypoints improves the AUC$@5^{\circ}$ by over 4\,pp.
But it requires careful weighting; otherwise, it causes a degraded performance ($\omega=1e-5$) or even a collapsed training ($\omega=1e-4$).
Yet, carefully tuned, it improves the quality of the heatmap, especially for low-resolution inputs, as can be seen in \cref{fig:heatmap_entropy}.
Please refer to the supplementary for additional ablations.

\section{Conclusion}
\label{sec:conclusion}

We presented a novel geometric reward for reinforcement learning–based training of sparse keypoint extractors under weak supervision.
 Importantly, this formulation fully exploits geometric consistency within a single image pair, eliminating the need for negative training pairs and further reducing the data requirements for RL-based keypoint learning.
Our experiments demonstrate that the resulting RIPE\texttt{++} extractor achieves competitive accuracy while relying on substantially weaker supervision than competing approaches. The simplified training setup also enables learning directly from video sequences. In particular, we show that training on endoscopic video streams is sufficient to obtain a specialized keypoint extractor for the medical domain, where geometric ground truth is often unavailable.

Beyond keypoint extraction, the positive-pair training paradigm also allows the construction of a reward signal for the matching stage.
Together, these results highlight the potential of reinforcement learning for jointly training keypoint extractors \emph{and} matchers under minimal supervision.

\section*{Acknowledgements}
This work was partly funded by the Federal Ministry of Education and Research (REFRAME, grant no. 01IS2407A) and the German Federal Ministry for Economic Affairs and Climate Action (DeepTrain, grant no. 19S23005D).

\bibliographystyle{splncs04}
\bibliography{main}

\newpage
{%
  \centering
  \Large\bfseries
  Supplementary Material for\\
  RIPE\texttt{++}: Reinforced Keypoint Learning from Positive Pairs Only\\[1em]
}

\section{Hyperparameters}

\Cref{tab:hyperparameter} documents our hyperparameters for the training of the keypoint extractor (left side) and the keypoint matcher (right side).

\begin{table}[htb]
\centering
\scriptsize
\caption{Overview of training hyperparameters for the keypoint extractor (left) and the matcher (right).}
\begin{tabular*}{\textwidth}{@{\extracolsep{\fill}}ccc@{\hskip 1.5em}ccc@{}}
\toprule
\multicolumn{3}{c}{Extractor} & \multicolumn{3}{c}{Matcher} \\
\cmidrule(r{0.0em}){1-3} \cmidrule(l{0.0em}){4-6}
Name & Value & Purpose & Name & Value & Purpose \\
\cmidrule(r{0.0em}){1-3} \cmidrule(l{0.0em}){4-6}
$\rho_{\text{in}}$  & 1.0   & reward \cref{eq:reward_pos}                               & $\nu_{\text{in}}$  & 1.0     & reward matcher training                                      \\
$\rho_{\text{out}}$ & -0.1  & penalty \cref{eq:reward_pos}                              & $\nu_{\text{out}}$ & -1.0    & penalty matcher training                                     \\
$\lambda$         & -1e-7 & penalty \cref{eq:reward_pos}                                & $\lambda$          & -1e-07  & penalty no match                                             \\
$\psi$              & 5.0   & weight $\mathcal{L}_{\text{desc}}$   & $\eta$             &  0.0001 & weight $\mathcal{L}_{\text{nm}}$ \cref{eq:final_loss_matcher}\\
$\omega$            & 1e-6  & weight $\mathcal{L}_{H}$             &                    &         &                                                              \\\bottomrule
\end{tabular*}
\label{tab:hyperparameter}
\end{table}

\section{Additional Architectural Experiments}
\label{sec:additional_improvements}

\subsection{Methods}

Beyond our core contributions, we explored several additional directions.
We document these investigations in the following, as they provide useful insights for practitioners working on RL-based keypoint learning.

\subsubsection{Distance-based Reward}

For further improvement of the reward signal, we replace the binary inlier/outlier reward with a distance-aware reward signal based on the Sampson distance~\cite{Hartley2004}, measuring how well a correspondence satisfies the epipolar constraint.
Given a putative match for the two keypoint locations $\mathbf{l}^A \in \mathbf{K}_A$ and $\mathbf{l}_B \in \mathbf{K}^B$ and the estimated fundamental matrix $\mathbf{F}$, the Sampson distance is defined as:

\begin{equation}
    d_S(\mathbf{l}_A, \mathbf{l}_B, \mathbf{F}) = \frac{({\mathbf{l}_B}^\top \mathbf{F} \mathbf{l}_A)^2}{(\mathbf{F}\mathbf{l}_A)_1^2 + (\mathbf{F}\mathbf{l}_A)_2^2 + (\mathbf{F}^\top\mathbf{l}_B)_1^2 + (\mathbf{F}^\top\mathbf{l}_B)_2^2},
\end{equation}
where $(\cdot)_i$ denotes the $i$-th component of a vector.
We convert this distance into a reward using a truncated $L_2$ kernel:
\begin{equation}
    r_{i,j}(d_S) = \begin{cases}
        1 - \frac{1}{2}\left(\frac{d_S}{\delta}\right)^2, & \text{if } d_S \leq \tau \\
        \rho_{\text{out}}, & \text{otherwise}
    \end{cases}
\end{equation}
where $\delta$ controls the curvature of the reward falloff, $\tau$ is the outlier threshold, and $\rho_{\text{out}}$ is again the penalty for outliers.

\begin{wrapfigure}{r}{0.5\textwidth}
    \centering
    \includegraphics[width=0.48\textwidth]{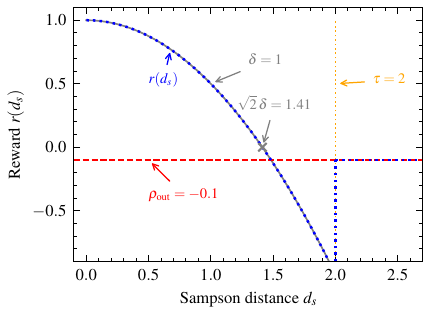}
    \vspace{-10pt}
    \caption{Illustration Distance-based Reward}
    \label{fig:L2-kernel}
    \vspace{-30pt}
\end{wrapfigure}

This formulation provides a smooth gradient for correspondences near the epipolar constraint, rather than treating all inliers equally, while explicitly penalizing geometric outliers.
Matches with zero Sampson distance receive maximal reward ($r=1$), with the reward decreasing quadratically as the distance increases, until the threshold $\tau$ is exceeded.
For the given values in \cref{fig:L2-kernel} and a RANSAC threshold of 1.5, inliers with $d_s>1.41$ already receive a (small) penalty, as they are deemed to be too far away from their respective epipolar line.
Outliers with $1.5 > d_s < 2.0$ receive an increasing penalty, as they were almost labeled inliers.
On the other hand, outliers with $d_s \geq 2.0$ receive a smaller constant penalty, as they are already sufficiently distant from deemed as inliers.

By replacing hard inlier decisions with a continuous geometric signal, the reward better reflects matching quality and improves gradient stability during training.

\subsubsection{Curriculum Learning}

Training keypoint detectors with weak supervision presents a challenge: early in training, the network produces many low-quality matches, resulting in noisy reward signals that can destabilize learning.
Following \cite{10.1109/cvpr52733.2024.00464}, we address this with a curriculum learning strategy that gradually increases training difficulty.

For each batch, we compute the reward $r_i$ for each sample $i$ and rank samples by their reward in descending order.
Only the top-$k\%$ of samples contribute to the loss; the remaining samples are masked out.
This ensures that the network initially learns from high-confidence matches where the reward signal is reliable.

The curriculum schedule follows a linear annealing:
\begin{equation}
    k(t) = \min\left(k_{\text{max}}, \, k_{\text{init}} + \left\lfloor \frac{t}{T_{\text{inc}}} \right\rfloor \cdot \Delta k \right),
\end{equation}
where $t$ is the current training step, $k_{\text{init}}$ is the initial percentage of samples used, $\Delta k$ is the increment step, $T_{\text{inc}}$ is the number of steps between increments, and $k_{\text{max}}$ is the maximum percentage.

In practice, we set $k_{\text{init}} = 30\%$, $\Delta k = 5\%$, $T_{\text{inc}} = 1000$, and $k_{\text{max}} = 80\%$.
This schedule allows the network to first establish reliable keypoint detection on easy samples before gradually incorporating harder cases with noisier supervision.
The upper bound $k_{\text{max}} < 100\%$ ensures that the most difficult samples -- which may contain erroneous rewards due to missing or too small visual overlap -- never dominate the training signal.

\begin{table}[tb]
\caption{Ablation of the different components on the MegaDepth1500 benchmark set with the longer side of the input rescaled to 1200 pixels and 2048 keypoints detected.}
\scriptsize
\begin{tabularx}{\textwidth}{CCCCCCCCCCC}
\toprule
\multicolumn{6}{c}{\textbf{Parameter}} & \multicolumn{4}{c}{\textbf{Metrics}} \\ 
\cmidrule(lr{.5em}){1-6} \cmidrule(lr{.5em}){7-11}
\rotatebox{90}{pos only}  & \rotatebox{90}{\parbox{2cm}{\raggedright contrastive}} & \rotatebox{90}{\parbox{2cm}{\raggedright infonce}} & \rotatebox{90}{\parbox{2cm}{\raggedright distance reward}} & \rotatebox{90}{curriculum} & \rotatebox{90}{entropy} & \rotatebox{90}{\parbox{2cm}{\raggedright \# RANSAC inlier}} & \rotatebox{90}{\parbox{2cm}{\raggedright \% RANSAC inlier}} & \rotatebox{90}{AUC$@5^{\circ}$} & \rotatebox{90}{AUC$@10^{\circ}$} & \rotatebox{90}{AUC$@20^{\circ}$} \\ \cmidrule(lr{.5em}){1-6} \cmidrule(lr{.5em}){7-11}
           &   {\cmark}   &                 &          &          &             &   297 &  34 & 51.83 & 65.37 & 75.94 \\
{\cmark}   &   {\cmark}   &                 &          &          &             &   427 &  41 & 52.42 & 66.78 & 78.03 \\ 
{\cmark}   &   {\cmark}   &                 & {\cmark} &          &             &   384 &  40 & 54.75 & 68.33 & 78.92 \\ 
{\cmark}   &   {\cmark}   &                 &          & {\cmark} &             &   347 &  38 & 52.57 & 65.84 & 76.43 \\ 
{\cmark}   &              &     {\cmark}    &          &          &             &   430 &  45 & 51.72 & 65.0  & 75.9  \\
{\cmark}   &              &     {\cmark}    & {\cmark} &          &             &   395 &  43 & 50.02 & 63.66 & 74.66 \\
{\cmark}   &              &     {\cmark}    &          & {\cmark} &             &   370 &  41 & 49.18 & 61.93 & 72.85 \\
{\cmark}   &              &     {\cmark}    & {\cmark} & {\cmark} &             &   362 &  40 & 51.04 & 64.29 & 74.6  \\

{\cmark}   &   {\cmark}   &                 & {\cmark} &          &  \num{1e-6} &   374 &  40 & 67.12 & 78.01 & 53.01 \\
\cmidrule(lr{.5em}){1-6} \cmidrule(lr{.5em}){7-11}
{\cmark}   &   {\cmark}   &                 &          &          &  \num{1e-6} &   366 &  41 & 69.44 & 79.18 & 56.59 \\\bottomrule
\end{tabularx}
\label{tab:ablations_supplementary}
\end{table}

\subsubsection{Improved Descriptor Loss}

RIPE~\cite{knzel2025ripe-7fa} used a hinge loss as auxiliary training signal to optimize the descriptors with
\begin{equation} 
    L_{\text{desc}}=
    \begin{dcases}
    \frac{1}{N}\sum_{n=1}^{N}\max(0,\mu+\delta_{+}^n-\delta_{h}^n), & \lambda_\kappa=1 \\
    \frac{1}{N}\sum_{n=1}^{N}\max(0,\mu - \delta_{+}^n), & \text{otherwise} 
    \end{dcases}
\label{eq:desc_loss}
\end{equation}
where $N$ is the number of inliers after the RANSAC filtering, $\mu$ is a positive threshold, and $\delta$ the L2-distance between two descriptors ($\delta_{+}^n$ being the L2-distance between two keypoint descriptors which got matched \textit{and} validated (i.e., identified as inliers) by geometric filtering while $\delta_{h}^n$ is the distance to the second closest neighbor).
In summary, the loss pulls putative matches closer in features space and repells matched descriptors for negative pairs, as their should be no matches ideally.

Despite having been proved effective the major drawback lies in the sparsity of the loss, as only the nearest and/~or the second nearest neighbor are taken into account.

To this end we draw inspiration from self-supervised works and aim to replace \cref{eq:desc_loss} with the InfoNCE~\cite{10.48550/arxiv.1807.03748} loss.

For two images $\mathbf{A}$ and $\mathbf{P}$ sets of descriptors $\mathbf{D}_A \in \mathbb{R}_{d \times N}$ and $\mathbf{D}_P \in \mathbb{R}_{d \times M}$ are extracted and L2-normalized.
We then calculate the similarity for every combination of descriptors with
\begin{equation}
    \mathbf{S} = \frac{1}{\tau}{\mathbf{D}_A}^T \mathbf{D}_P \in \mathbb{R}^{N \times M}
\end{equation}
with $\tau$ being a temperature hyperparameter that controls the sharpness of the resulting distribution.
Applying \textsc{softmax} on each row of $\mathbf{S}$ with 
\begin{equation}
p_{ij}=\frac{e^{s_{ij}}}{\sum_{k=i}^M e^{s_{ik}}},
\label{eq:softmax}
\end{equation}
effectively results in the probabilities of some descriptor $d^a_i$ matching some descriptor $d^b_j$.

With $k^i$ giving the putative correct match (from NN-matching and RANSAC filtering, not groundtruth) of descriptor $d_i$ we formulate the final cross-entropy loss 
\begin{equation}
    L_i = - \frac{1}{N}\sum_{i=1}^N \log p_{i,k^i}
\end{equation}
effectively enforcing the probability mass to concentrate on the correct match.

\subsubsection{Evaluation}

The ablation in \cref{tab:ablations_supplementary} confirms that training on positive pairs only consistently improves results.
Replacing the descriptor loss, however, does not: while the number of RANSAC inliers increases notably, this does not translate into improved pose accuracy.
Curriculum learning, entropy regularization, and the distance-based reward each individually improve results, though their combination does not yield further cumulative gains.
The best overall configuration uses positive-only training, entropy regularization, and the contrastive descriptor loss.

\section{Influence of Removing Negative Samples}

\begin{figure}[htb]
    \centering
    \includegraphics[width=1.0\linewidth]{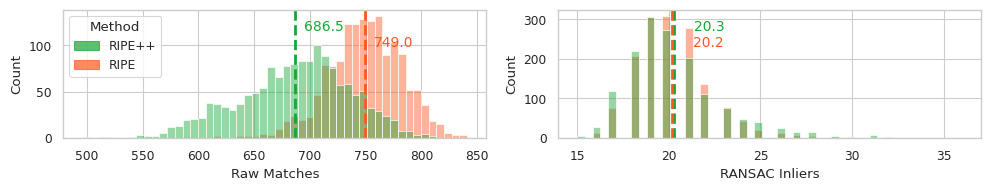}
    \caption{Number of false NN matches (left) and RANSAC inliers (right) for negative image pairs.}
    \label{fig:num_matches}
\end{figure}

While removing the negative pairs from RIPE\texttt{++}'s training proved beneficial in our evaluations, we examine whether a model trained without negative pairs produces more spurious correspondences on geometrically unrelated image pairs.
To this end, we sample 1500 negative pairs from the MegaDepth-1500 test set by pairing images from two different scenes, which therefore share no common underlying 3D geometry.
We extract keypoints and match them using the same settings as in Sec. 4.1, and visualize in \cref{fig:num_matches} the histograms of false nearest-neighbor matches and of RANSAC inliers after filtering with the fundamental matrix. 
The results clearly show that the absence of negative pairs does not increase the number of false matches; in fact, it decreases them by \SI{8.4}{\percent}, indicating that the negative pairs are not necessary to suppress spurious correspondences.

\section{Derivation of Weakly-Supervised LightGlue Training}
\label{sec:disk_lightglue}

We provide a detailed derivation of the connection between the DISK~\cite{Tyszkiewicz.2020} 
and LightGlue~\cite{lindenberger_2023_lightglue} formulations.

\subsubsection{Background: LightGlue}

Given two sets of local features from images $A$ and $B$, each with normalized 2D positions 
$\mathbf{p}_i$ and descriptors $\mathbf{d}_i \in \mathbb{R}^d$, LightGlue processes point 
representations through $L$ stacked self- and cross-attention layers.
After each layer, a classifier determines whether inference can be halted early to reduce computation.
A lightweight head then computes pairwise similarity scores
\begin{equation}
    S_{i,j} = \text{Linear}\left(\mathbf{x}_i^A\right)^\top \text{Linear}\left(\mathbf{x}_j^B\right),
    \label{eq:appendix_sim_lightglue}
\end{equation}
and per-point matchability scores
\begin{equation}
    \sigma_i = \text{Sigmoid}\left(\text{Linear}\left(\mathbf{x}_i\right)\right) \in [0,1],
    \label{eq:appendix_matchability_lightglue}
\end{equation}
encoding the likelihood that point $i$ has a corresponding point in the other image.
Both terms are combined into a soft partial assignment matrix
\begin{equation}
    P_{i,j} = \sigma_i^{A} \, \sigma_j^{B} \,
    \underset{k \in \mathcal{A}}{\text{Softmax}}(\mathbf{S}_{k,j})_i \;
    \underset{k \in \mathcal{B}}{\text{Softmax}}(\mathbf{S}_{i,k})_j.
    \label{eq:appendix_partial_assignment_lightglue}
\end{equation}
The network is trained by minimizing the negative log-likelihood of ground-truth assignments 
at each layer $\ell$:
\begin{equation}
\begin{split}
    \mathcal{L} = -\frac{1}{L} \sum_{\ell} \Bigg(
    &\frac{1}{|\mathcal{M}|} \sum_{(i,j) \in \mathcal{M}} \log\, {}^\ell\! P_{i,j} \\
    &+ \frac{1}{2|\bar{\mathcal{A}}|} \sum_{i \in \bar{\mathcal{A}}} \log \left(1 - {}^\ell\sigma_{i}^{A}\right) \\
    &+ \frac{1}{2|\bar{\mathcal{B}}|} \sum_{j \in \bar{\mathcal{B}}} \log \left(1 - {}^\ell\sigma_{j}^{B}\right)
    \Bigg),
    \label{eq:loss_lightglue}
\end{split}
\end{equation}
where $\mathcal{M}$ denotes ground-truth matches, and $\bar{\mathcal{A}} \subseteq \mathcal{A}$, 
$\bar{\mathcal{B}} \subseteq \mathcal{B}$ are the unmatched points in each image.

\subsubsection{Background: DISK Policy Gradient}

DISK~\cite{Tyszkiewicz.2020} formulates keypoint detection and matching as a reinforcement 
learning problem, optimizing the expected reward over sampled matches rather than a supervised loss.
A keypoint at location $p$ within patch $u$ is selected and accepted with probability
\begin{equation}
    P(p \mid \mathbf{K}^u) = \text{Softmax}(\mathbf{K}^u)_p \cdot \text{Sigmoid}(\mathbf{K}^u_p),
    \label{eq:appendix_prob_selection_disk}
\end{equation}
where $\mathbf{K}^u_p$ is the detection heatmap logit at location $p$.
Match probabilities are derived from the L2-distance matrix $\mathbf{d}$ between descriptors, 
where rows and columns define categorical distributions for the forward ($A \to B$) and reverse 
($B \to A$) directions:
\begin{equation}
    P_{A \to B}(j \mid \mathbf{d}, i) = \text{Softmax}(-\theta_M \, \mathbf{d}(i,\cdot))_j,
\end{equation}
with $\theta_M$ the inverse softmax temperature. The negative sign converts L2 distance into 
affinity, analogous to how LightGlue uses the dot product directly (where zero denotes 
perpendicular, rather than identical, descriptors).

The policy gradient then takes the form
\begin{equation}
\begin{split}
    \nabla_\theta \, \underset{M_{A \leftrightarrow B}}{\mathbb{E}} R(M_{A \leftrightarrow B})
    &= \underset{F_A, F_B}{\mathbb{E}} \sum_{i,j} \Big[
    P(i,j \mid F_A, F_B) \cdot r(i,j) \cdot \nabla_{\theta} \Gamma_{ij}
    \Big], \\
    \Gamma_{ij} &= \log P(i,j \mid F_A, F_B)
    + \log P(F_{A,i} \mid A)
    + \log P(F_{B,j} \mid B),
    \label{eq:appendix_disk_gradient}
\end{split}
\end{equation}
where the sum runs only over matched pairs, since unmatched keypoints receive zero reward.
Crucially, since all match probabilities $P(i,j)$ can be computed in closed form, the inner 
expectation is evaluated exactly -- no sampling is required and no variance is introduced at 
this level. Variance arises only from the empirical approximation of the outer expectation 
over feature sets $F_A, F_B$.

\subsubsection{Adapting DISK's Policy Gradient to LightGlue}

The structural parallel between DISK and LightGlue is direct: both decompose a soft assignment 
into a \emph{matchability/ keypointness} term and a \emph{match} term.
We therefore substitute DISK's keypoint selection probability 
(\cref{eq:appendix_prob_selection_disk}) with LightGlue's matchability score 
(\cref{eq:appendix_matchability_lightglue}):
\begin{equation}
    P_i(\mathbf{x}_i \mid I) = \text{Sigmoid}\left(\text{Linear}\left(\mathbf{x}_i\right)\right),
\end{equation}
and replace the descriptor-distance-based match probability with LightGlue's attention-based 
formulation:
\begin{equation}
    P(i,j \mid A, B) =
    \underset{k \in \mathcal{A}}{\text{Softmax}}(\mathbf{S}_{k,j})_i \;
    \underset{k \in \mathcal{B}}{\text{Softmax}}(\mathbf{S}_{i,k})_j.
    \label{eq:appendix_prob_match_matcher}
\end{equation}
Substituting into \cref{eq:appendix_disk_gradient} yields the gradient estimator used in our 
training:
\begin{equation}
\begin{split}
    \nabla_\theta \, \underset{M_{A \leftrightarrow B}}{\mathbb{E}} R(M_{A \leftrightarrow B})
    &= \underset{F_A, F_B}{\mathbb{E}} \sum_{i,j} \Big[
    P(i,j \mid A, B) \cdot r(i,j) \cdot \nabla_{\theta} \Gamma_{ij}
    \Big], \\
    \Gamma_{ij} &= \log P(i,j \mid A, B)
    + \log P_i(\mathbf{x}_i \mid A)
    + \log P_j(\mathbf{x}_j \mid B).
    \label{eq:appendix_ripe_match_gradient}
\end{split}
\end{equation}

\begin{table}[htb]
\label{tab:aachen_v1_1}
\caption{Results on outdoor visual localization using Aachen Day-Night v1.1.}
\small
\centering
\begin{tabularx}{\textwidth}{CCCCCCC}
\toprule
\multicolumn{1}{c}{\multirow{2}{*}{Method}} & \multicolumn{3}{c}{Day}  & \multicolumn{3}{c}{Night} \\ \cmidrule(l){2-4} \cmidrule(l){5-7} 
\multicolumn{1}{c}{}  & .25m/\ang{2}     & .5m/\ang{5}        & 5m/\ang{10}           & .25m/\ang{2}         & .5m/\ang{5}        & 5m/\ang{10} \\ \midrule
ALIKED                & 85.5             & 92.0               & 95.8                  & 66.0                 & 80.6               & 92.7  \\
DaD                   & 88.9             & 93.2               & 97.0                  & 69.6                 & 85.9               & 96.9  \\
DeDoDe                & 81.8             & 89.1               & 93.4                  & 55.5                 & 68.1               & 78.0  \\ 
DISK                  & 81.9             & 91.4               & 95.4                  & 61.8                 & 75.9               & 86.9  \\ 
RaCo                  & 87.4             & 93.4               & 97.1                  & 72.8                 & 86.9               & 96.9  \\ 
RIPE                  & 80.9             & 88.5               & 92.8                  & 45.0                 & 60.2               & 74.9  \\ \midrule
\textbf{RIPE\texttt{++}} & 81.3             & 89.0               & 93.7               & 54.5                 & 67.5               & 80.1  \\ 
$\downarrow$\scriptsize + Tokyo & \rss{-2.5} & \rss{-1.9}        & \rss{-1.5}         & \gss{+5.7}           & \gss{+4.8}         & \gss{+3.7} \\
\textbf{RIPE\texttt{++}}       & 78.8             & 87.1               & 92.2                  & 60.2                 & 72.3               & 83.8 \\ \bottomrule
\end{tabularx}
\end{table}

\begin{figure}[b]
    \centering
    \begin{subfigure}[b]{\textwidth}
            \includegraphics[width=0.32\textwidth]{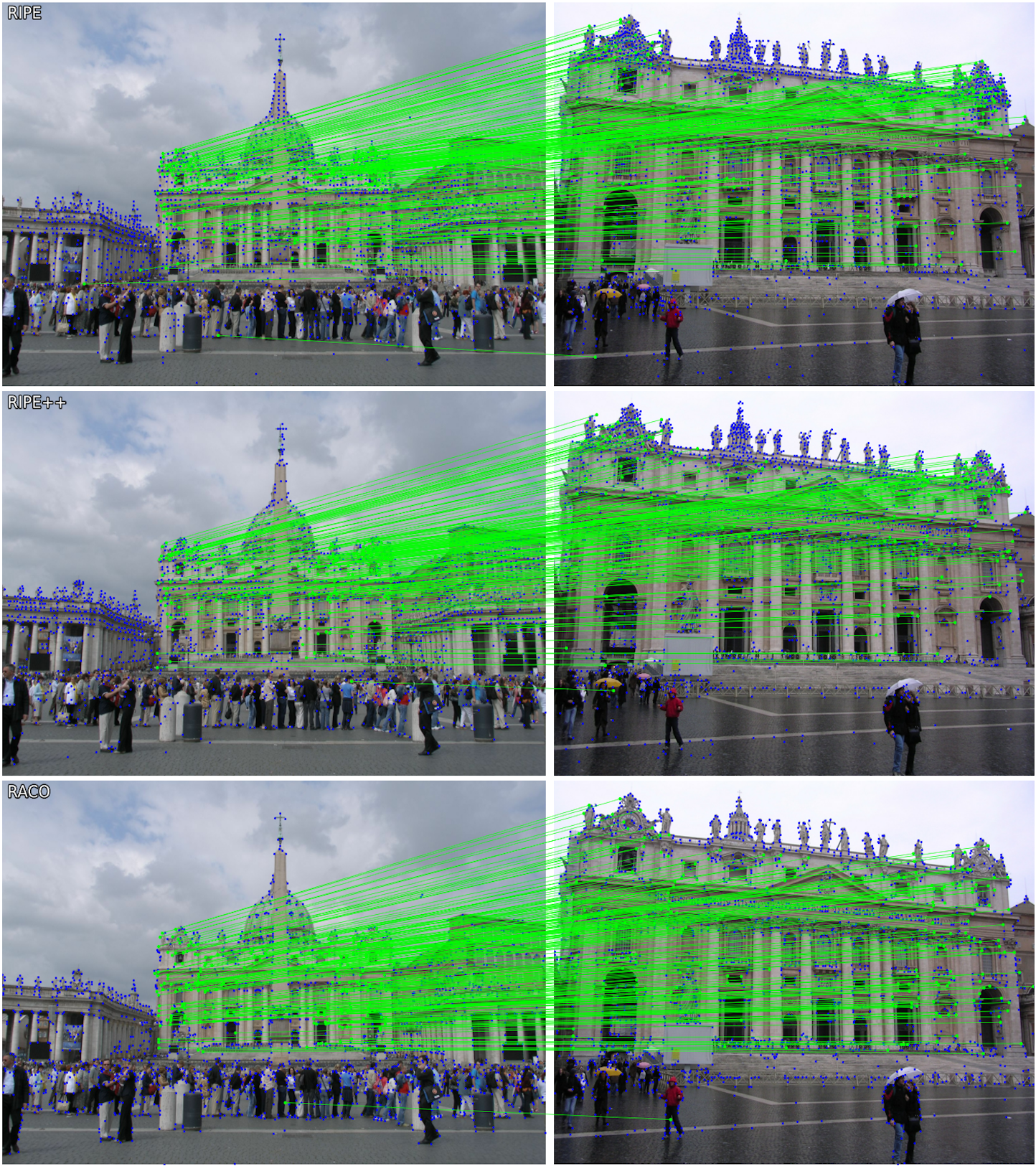}
            \includegraphics[width=0.32\textwidth]{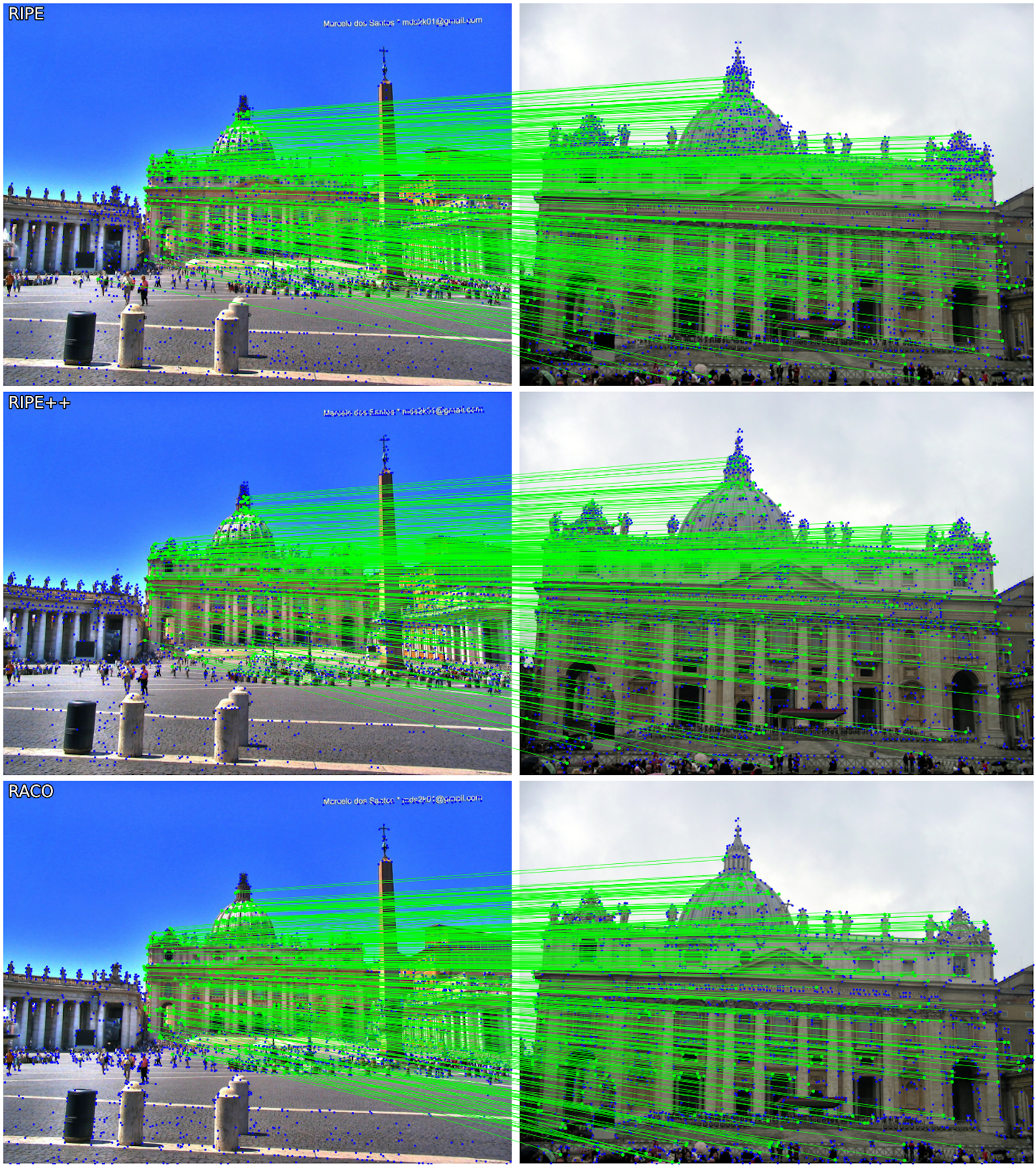}
            \includegraphics[width=0.32\textwidth]{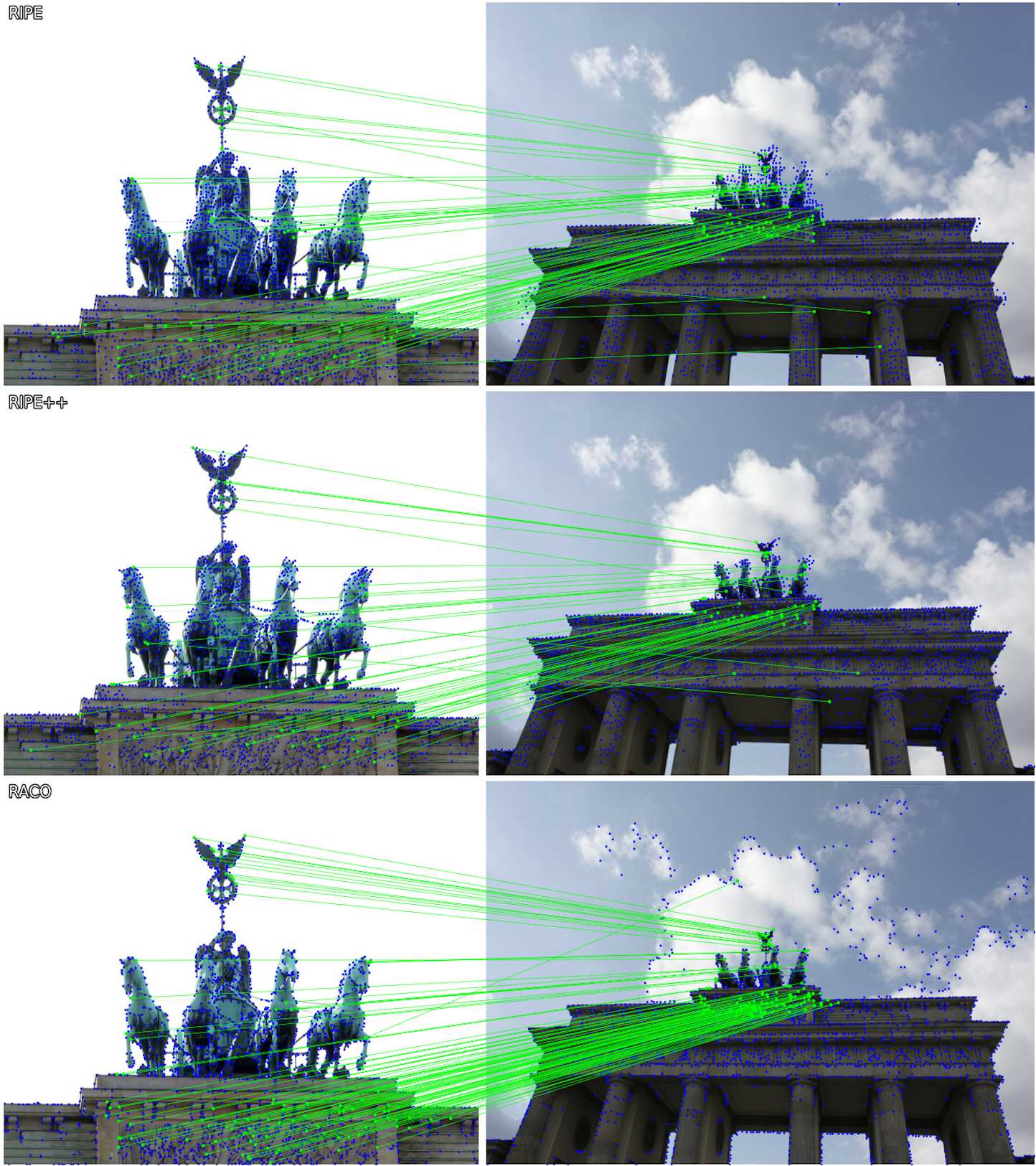}
    \end{subfigure}
\caption{Qualitative results on the MegaDepth1500 benchmark dataset for, RIPE~\cite{knzel2025ripe-7fa}~(top), RIPE\texttt{++}~(ours, middle) and  RaCo~\cite{Shenoi2026}~(bottom).}
\label{fig:qualitative_results_megadepth}
\end{figure}

\section{Outdoor localization day-night}
\label{sec:aachen_v1_1}

\PAR{Dataset} We evaluate our approach on visual localization, \ie the estimation of the 6-DoF pose of a query image relative to a 3D scene model, using the Aachen~v1.1 dataset~\cite{Zhang2020ARXIV,Sattler2018CVPR,Sattler2012BMVC}.
This dataset is particularly challenging due to large viewpoint and illumination changes, including day-to-night transitions.

\PAR{Metrics}
Using the HLoc framework~\cite{10.1109/cvpr.2019.01300} (commit c13273b with pycolmap 3.13.0), we first triangulate a 3D model from the 6,697 reference images, then retrieve 50 candidate images per query via NetVLAD~\cite{Arandjelovic.2016} and match them against each of the 1,015 queries (824 daytime, 191 nighttime).
Camera poses are estimated using a Perspective-n-Point solver with RANSAC, and we report AUC at thresholds of 0.25\,\si{\m}/\ang{2}, 0.5\,\si{\m}/\ang{5}, and 1.0\,\si{\m}/\ang{10}.

\PAR{Baselines}
We compare RIPE\texttt{++} against state-of-the-art methods for sparse keypoint extraction.
Since RaCo and DaD are pure detectors, we pair both with ALIKED-n16 descriptors for evaluation.
Images are resized to 1600 pixels on the longer side, and all methods are evaluated using mutual nearest-neighbor matching to isolate the quality of the extracted keypoints.

\PAR{Results}
\Cref{tab:aachen_v1_1} shows results on the Aachen~v1.1 benchmark.
RIPE\texttt{++} consistently improves over RIPE across all thresholds, with particularly notable gains under nighttime conditions (+9.5\,pp at 0.25\,\si{\m}/\ang{2}, +7.3\,pp at 0.5\,\si{\m}/\ang{5}, +5.2\,pp at 5\,\si{\m}/\ang{10}).
On daytime queries, RIPE\texttt{++} performs on par with DeDoDe and DISK.
A gap to fully-supervised methods such as DaD and RaCo remains, which is expected given that RIPE\texttt{++} trains without ground-truth depth or pose and, unlike RaCo, without synthetic homography augmentation.
The authors of RIPE~\cite{knzel2025ripe-7fa} also showed that the weakly-supervised training scheme facilitates the integration of additional training data. 
To verify that this still holds for RIPE\texttt{++}, we replace \SI{20}{\percent} of the training data with day/night pairs from the query images of the Tokyo~24/7~\cite{Torii.2015} dataset.
As \cref{tab:aachen_v1_1} shows, this yields a slight degradation in the daytime setting—expected, given the reduced number of remaining daytime pairs—but a clear improvement for nighttime localization.

\begin{figure}[t]
    \centering
    \begin{subfigure}[b]{\textwidth}
            \includegraphics[width=0.32\textwidth]{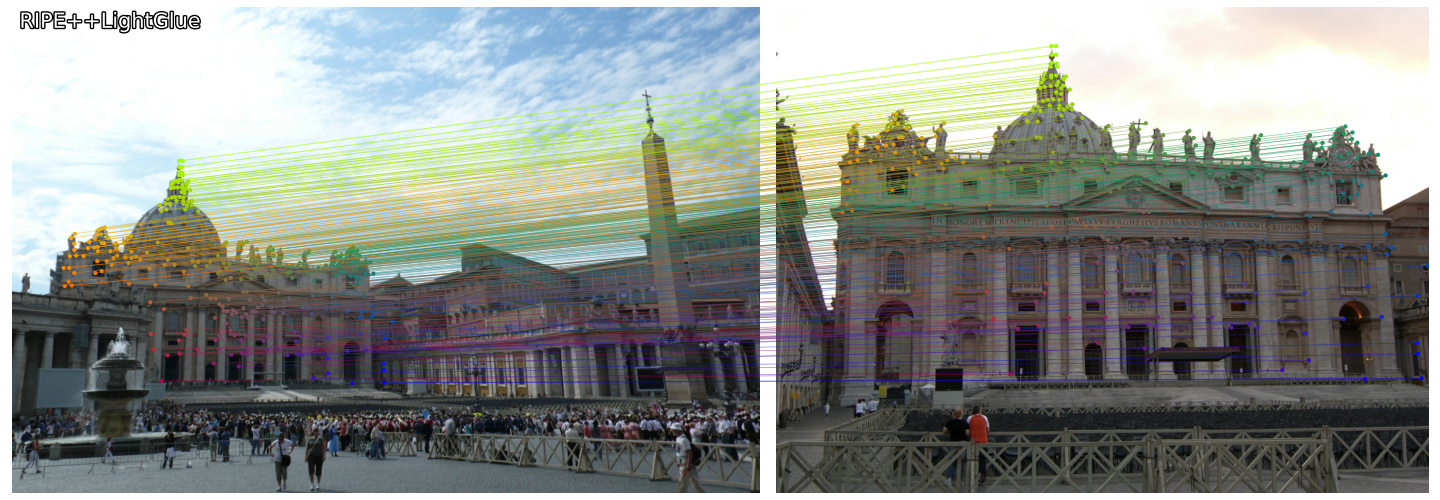}
            \includegraphics[width=0.32\textwidth]{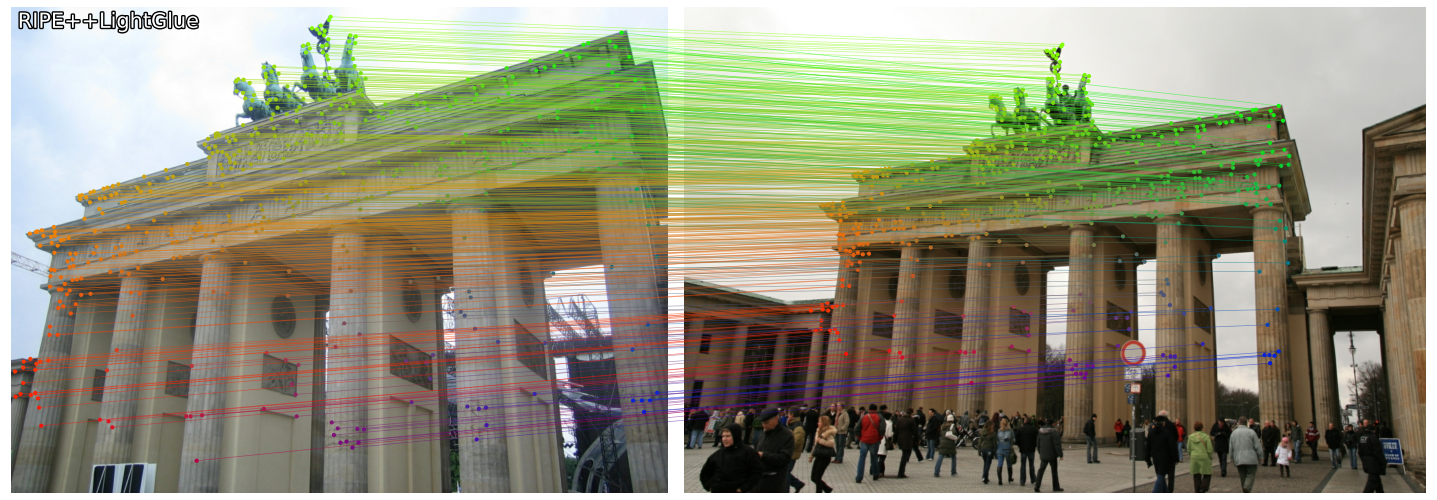}
            \includegraphics[width=0.32\textwidth]{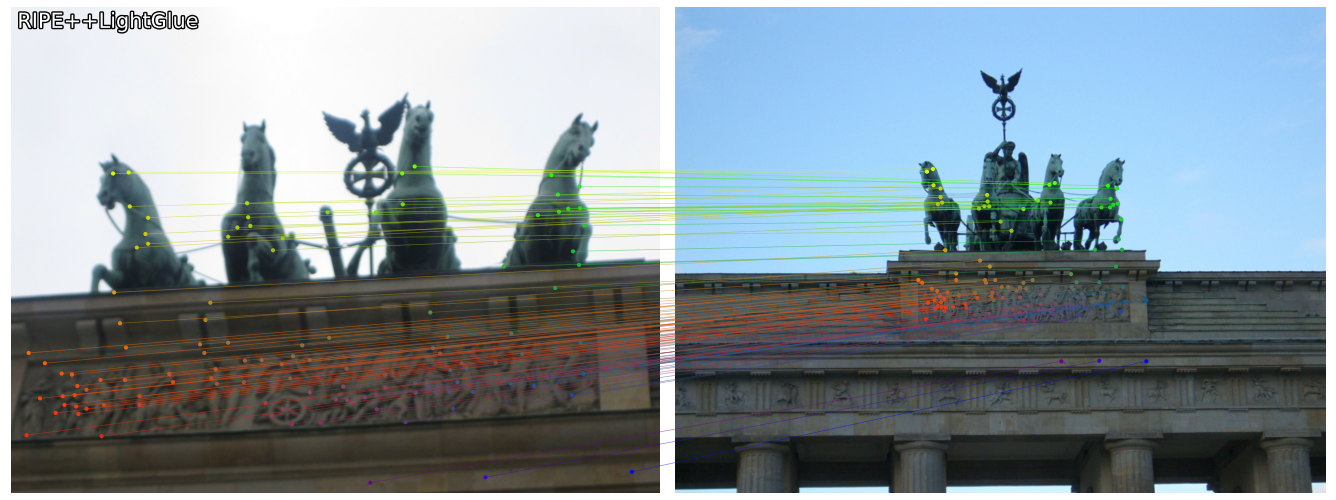}
    \end{subfigure}
\caption{Qualitative results on MegaDepth1500 for RIPE\texttt{++} paired with our weakly-supervised LightGlue matcher.}
\label{fig:qualitative_results_megadepth_LG}
\end{figure}

\begin{figure}[h]
    \centering
    \begin{subfigure}[b]{\textwidth}
            \includegraphics[width=0.32\textwidth]{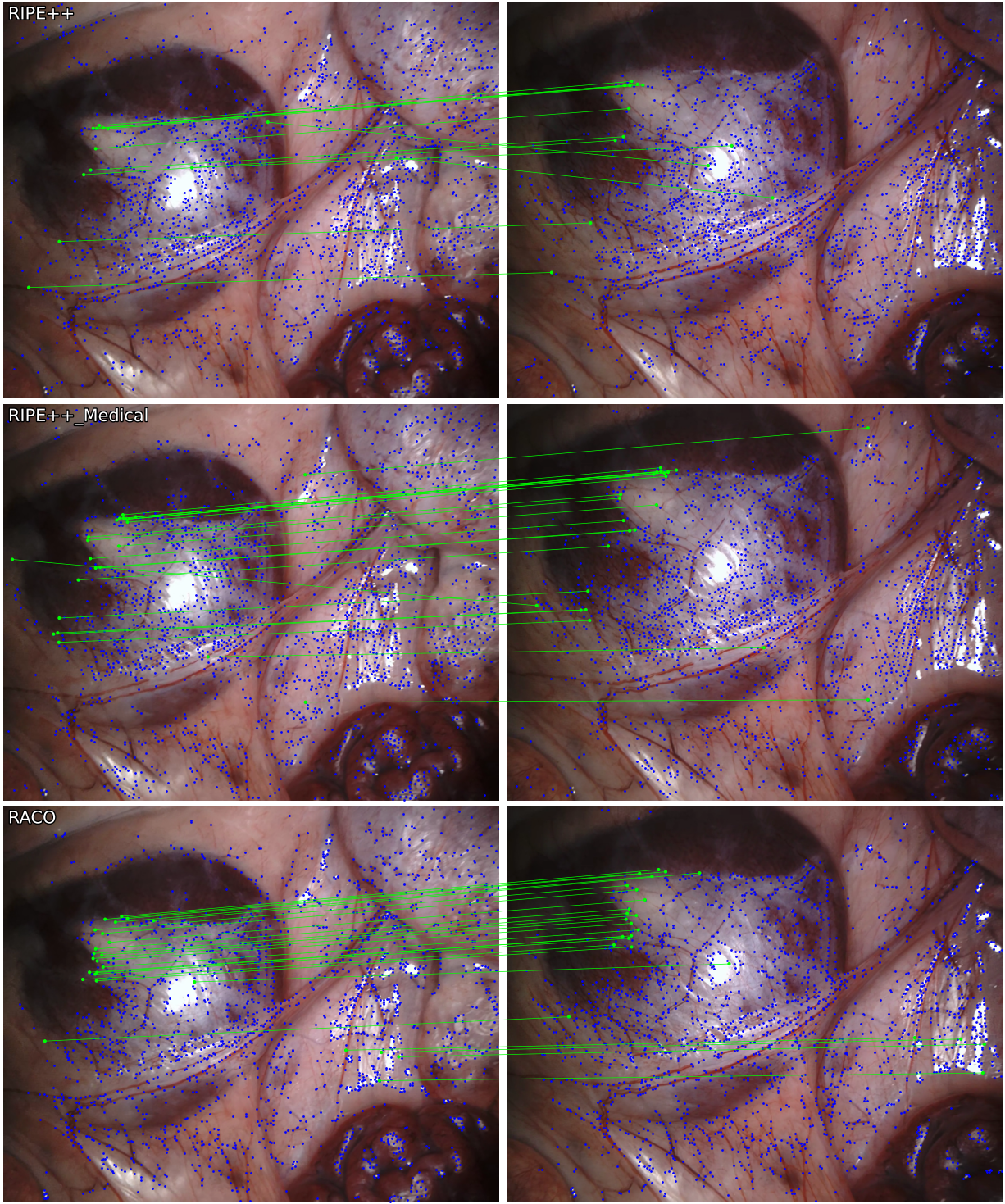}
            \includegraphics[width=0.32\textwidth]{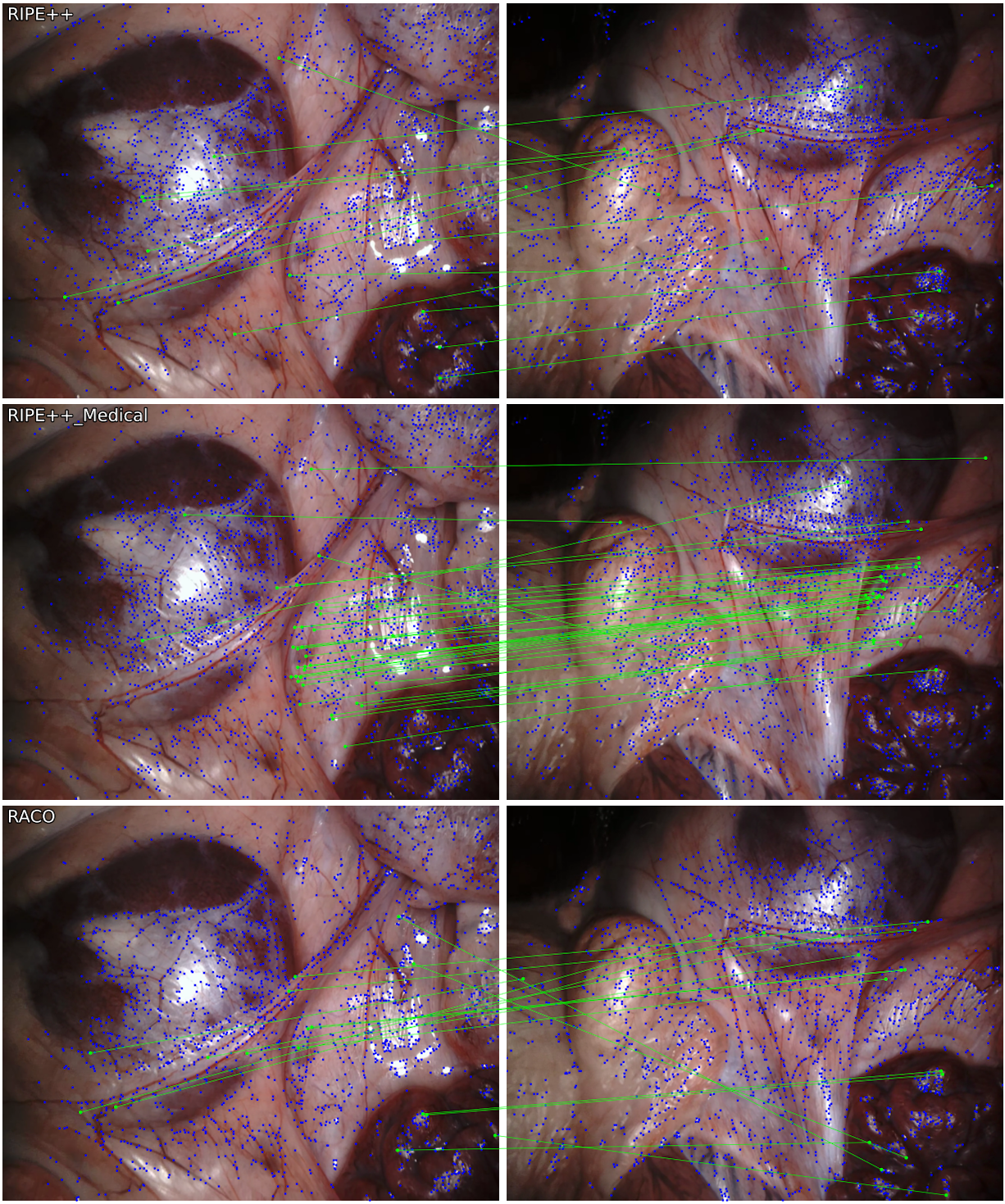}
            \includegraphics[width=0.32\textwidth]{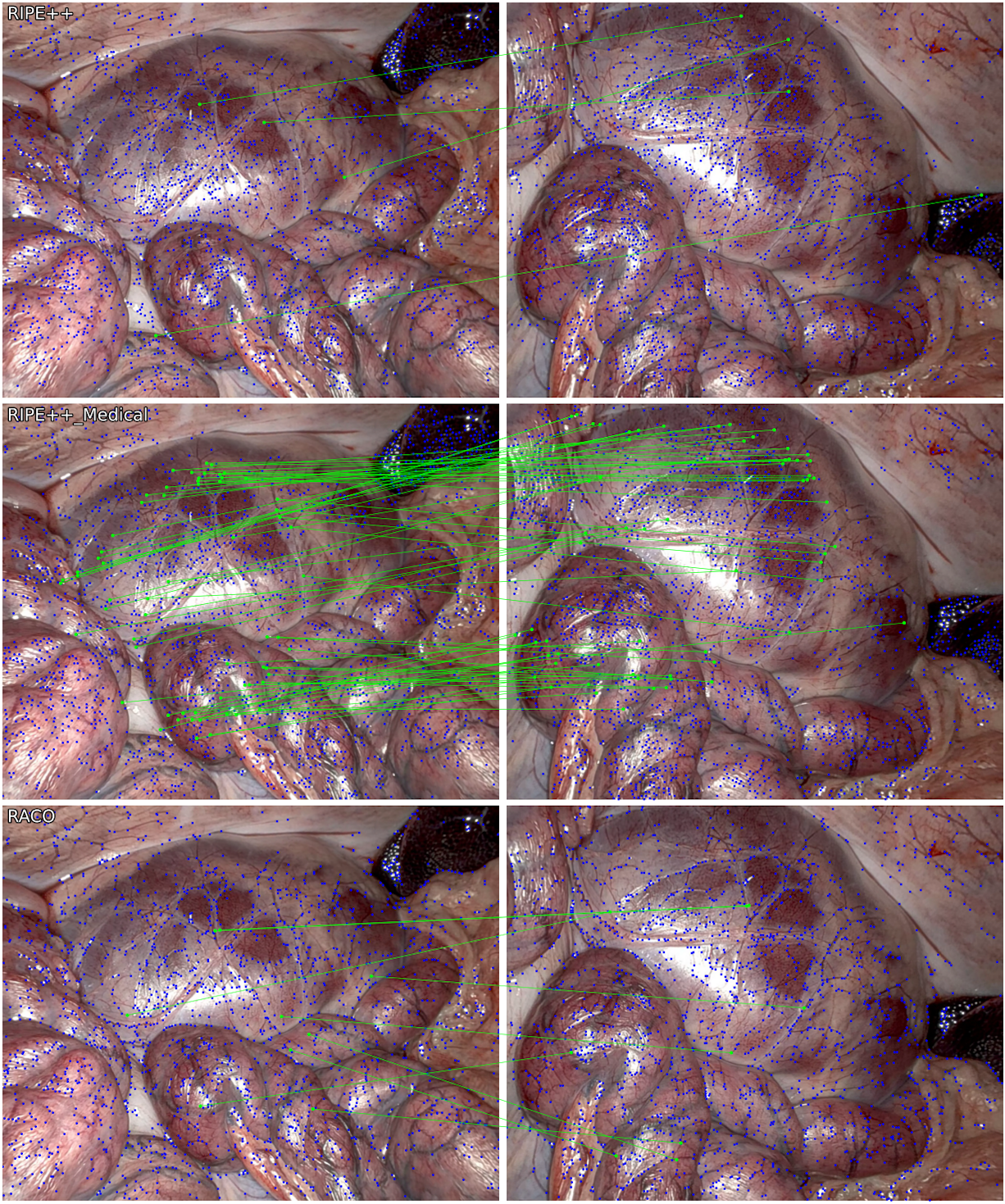}
    \end{subfigure}
\caption{Qualitative results on the Scared1500 benchmark dataset for RIPE\texttt{++}~(ours, top), RIPE\texttt{++}~Medical~(ours, middle) and RaCo~\cite{Shenoi2026}~(bottom).}
\label{fig:qualitative_results_scared}
\end{figure}

\section{Qualitative Results}

We show qualitative results on MegaDepth1500 for sparse keypoint extraction (\cref{fig:qualitative_results_megadepth}), RIPE\texttt{++} in combination with LightGlue (\cref{fig:qualitative_results_megadepth_LG}) and on our proposed SCARED1500 benchmark (\cref{fig:qualitative_results_scared}), using the same settings as in the respective quantitative evaluations.
\Cref{fig:qualitative_results_megadepth} shows that both RIPE\texttt{++} and RIPE learn to suppress keypoint detection in uninformative regions such as sky, while RaCo~\cite{Shenoi2026} occasionally detects keypoints on clouds (\eg, the right-most image), likely due to its reliance on synthetic homographies during training.
Compared to RIPE, our method tends to localize keypoints more precisely at object boundaries, as visible around the \textit{Quadriga} in the right-most image.
\Cref{fig:qualitative_results_scared} illustrates the substantial domain gap between the typical MegaDepth training distribution and the medical domain, and demonstrates how training RIPE\texttt{++} on medical data yields both more numerous and more accurate matches.

\end{document}